\documentclass{article}

\usepackage{PRIMEarxiv}

\usepackage[utf8]{inputenc} 
\usepackage[T1]{fontenc}    
\usepackage{hyperref}       
\usepackage{url}            
\usepackage{booktabs}       
\usepackage{amsfonts}       
\usepackage{nicefrac}       
\usepackage{microtype}      
\usepackage{lipsum}
\usepackage{fancyhdr}       
\usepackage{graphicx}       
\graphicspath{{media/}}     

\usepackage{amsmath}
\usepackage{makecell}

\usepackage{caption}
\usepackage{subcaption}
\usepackage{multirow}
\usepackage{graphicx}
\usepackage{tikz}
\usepackage{natbib}

\usepackage{adjustbox}
\usepackage{multirow}
\usepackage{lineno}
\usepackage[most]{tcolorbox}
\usepackage{xcolor}
\usepackage{listings}
\usepackage{enumitem}
\usepackage{subcaption}
\usepackage{graphicx}
\definecolor{darkblue}{rgb}{0, 0, 0.5}
\hypersetup{colorlinks=true, citecolor=darkblue, linkcolor=darkblue, urlcolor=darkblue}
\definecolor{promptcolor}{RGB}{245, 245, 250} 
\newtcolorbox{promptbox}{
  colback=promptcolor,
  colframe=blue!50!black,
  boxrule=0.5pt,
  arc=2pt,
  left=4pt,
  right=4pt,
  top=4pt,
  bottom=4pt,
  width=\columnwidth, 
  before skip=8pt,
  after skip=8pt,
}

\title{Can LLMs Cook Jamaican Couscous? A Study of Cultural Novelty in Recipe Generation
}

\author{
  Florian Carichon \\
  MILA, McGill University \\
  Montréal, Canada\\
  \texttt{florian.carichon@mila.quebec} \\
   \And
  Romain Rampa \\
  École de technologie supérieure (ÉTS) \\
  Montréal, Canada\\
  \texttt{romain.rampa@etsmtl.ca} \\
  \And 
  Golnoosh Farnadi \\
  MILA, McGill University \\
  Montréal, Canada\\
  \texttt{farnadig@mila.quebec}
}

\begin{document}
\maketitle

\begin{abstract}
Large Language Models (LLMs) are increasingly used to generate and shape cultural content, ranging from narrative writing to artistic production. While these models demonstrate impressive fluency and generative capacity, prior work has shown that they also exhibit systematic cultural biases, raising concerns about stereotyping, homogenization, and the erasure of culturally specific forms of expression. Understanding whether LLMs can meaningfully align with diverse cultures beyond the dominant ones remains a critical challenge. In this paper, we study cultural adaptation in LLMs through the lens of cooking recipes, a domain in which culture, tradition, and creativity are tightly intertwined. We build on the \textit{GlobalFusion} dataset, which pairs human recipes from different countries according to established measures of cultural distance. Using the same country pairs, we generate culturally adapted recipes with multiple LLMs, enabling a direct comparison between human and LLM behavior in cross-cultural content creation. Our analysis shows that LLMs fail to produce culturally representative adaptations. Unlike humans, the divergence of their generated recipes does not correlate with cultural distance. We further provide explanations for this gap. We show that cultural information is weakly preserved in internal model representations, that models inflate novelty in their production by misunderstanding notions such as creativity and tradition, and that they fail to identify adaptation with its associated countries and to ground it in culturally salient elements such as ingredients. These findings highlight fundamental limitations of current LLMs for culturally oriented generation and have important implications for their use in culturally sensitive applications.
\end{abstract}

\section{Introduction}

Large Language Models (LLMs) are increasingly used to produce cultural content\footnote{Culture defines common ground knowledge, aboutness, and the values, norms, and beliefs of a group of people \citep{paletz2008implicit, hershcovich2022challenges}. Cultural is then employed to characterize an object through the lens of these norms and values for a defined group of people. For example, cultural content is used to determine content encompassing ideas, values, and artistic expressions (like film, music, literature, and art), and cultural creation reflects how new ideas and forms of expression reflect the norms and values of a population \citep{boden2010creativity}.}, ranging from creative and narrative storytelling \citep{xie2023next} to art and music production \citep{liu2025wavjourney}, and even to creating new objects \citep{meincke2024using}. As generative models participate in creative processes, there is a growing need to assess whether they contribute to culturally meaningful creation rather than merely producing novel or fluent outputs \citep{brinkmann2023machine}. However, recent work has shown that LLMs exhibit systematic cultural biases. Their outputs tend to align disproportionately with the values of Western societies \citep{johnson2022ghost}, while struggling to represent diverse cultural value systems faithfully \citep{masoud2025cultural, atari2023humans}. These limitations can lead to stereotyping, cultural flattening, or erasure of culturally specific distinctions \citep{yu2025entangled}. As a result, understanding how LLMs encode, preserve, or transform cultural information has become a central concern in NLP research.

Most existing evaluations of cultural alignment rely on question–answering or survey-style assessments, in which models are explicitly prompted to express preferences, values, or beliefs. While useful, such approaches are inherently fragile: slight variations in prompt phrasing can lead to substantial changes in measured cultural alignment \citep{khan2025randomness} that do not align with the culturally aligned human counterparts \citep{arora2023probing}. These limitations have motivated a shift toward generative evaluations, which analyze free-form LLM outputs and assess cultural behavior indirectly \citep{adilazuarda2024towards, tao2024cultural, cao2023assessing}. However, evaluation methods typically focus on a limited set of dimensions and compare only a small number of cultural contexts, often treated in isolation \citep{hershcovich2022challenges, pawar2025survey} (i.e. assessing LLMs on only Arabic perception of one dimension at a time like traditional values or political views), without focusing on the cross-cultural representations at the intersection of complex value systems as with creative cultural products.

A core notion underlying cross-cultural analysis is cultural difference, often conceptualized in terms of cultural distance. Cultural distance has been widely studied in the social sciences using a variety of indicators, including geography, language, religion, and value systems \citep{shenkar2001cultural, hofstede2013vsm, WVSCultMap}. As traditions and perceptions of authenticity maintain group and cultural identity \citep{hobsbawm2012invention}, they shape how individuals perceive novelty, relevance, and innovation \citep{goh2009culture, kim2013information, paletz2008implicit}. As creativity can be defined as the ability to create cultural artifacts considered as new, surprising, and valuable \citep{boden2010creativity}, it operates as a mechanism for re-expressing traditions and meanings in the form of new products \citep{appadurai1988make}. This framework then allows for analyzing how LLMs encode specificity and originality that distinguish cultures and values.

While recent research on LLM creativity has primarily focused on output diversity, novelty, and the types or limits of creativity LLMs can exhibit \citep{peeperkorn2024temperature, franceschelli2025creativity}, this line of work has largely abstracted away from cultural grounding. In parallel, studies on cultural alignment in LLMs have concentrated on factual or evaluative tasks rather than generative tasks \citep{hu2024bridging, cao2024cultural}, especially overlooking models’ ability to generate culturally meaningful and creative adaptations. As a result, the intersection between creativity and cultural representativity in generative model outputs remains underexplored. A first step in this direction was introduced by \cite{carichon2025crossing}, who operationalized cultural creativity through measures of novelty and divergence in cooking recipes. The culinary domain is particularly suitable for such analysis, as cuisine has long been recognized as a central carrier of cultural identity, tradition, and authenticity \citep{appadurai1988make}, and culinary adaptation across cultures typically involves nuanced transformations rather than radical invention \citep{guerrero2009consumer}. We extend this line of work by proposing a benchmark that extends the \textit{GlobalFusion} dataset \citep{carichon2025crossing} by generating analogous recipe adaptations with LLMs and comparing their divergence with human adaptations. In this article, our contributions are the following:
\begin{itemize}
\item We introduce a novel framework and benchmark to produce LLM-generated cooking recipes, enabling the study of generative cultural artifacts across countries.
\item We evaluate whether LLMs produce cultural adaptations whose novelty and divergence correlate with cultural distance and compare these patterns to human adaptations.
\item We analyze the extent to which LLMs reflect culturally grounded creativity and traditions, as opposed to surface-level or stylistic variation.
\item We conduct a detailed analysis of LLM-generated recipes, combining internal representations, country identification, and ingredient-level grounding, to provide insights into how LLMs encode culture.
\end{itemize}

\section{Related Work}

\subsection{LLMs cultural bias and alignment}

Substantial prior work has highlighted that LLMs encode cultural biases reflecting the values of their training data \citep{hershcovich2022challenges,pawar2025survey}. A first type of studies relies on adapting survey-based methodologies in which models are asked to select from predefined options to reveal their preferences and values \citep{mazeika2025utility, adilazuarda2024towards}. Another line of work focuses on generative evaluations, analyzing free-form outputs in a manner closer to qualitative methods in social sciences \citep{adilazuarda2024towards,pawar2025survey}. Overall, these studies show that LLMs tend to reflect values aligned with Western, Educated, Industrialized, Rich, and Democratic societies \citep{johnson2022ghost}, resembling those of specific cultures, such as Dutch cultural values \citep{atari2023humans}. Similarly, \cite{masoud2025cultural}, using Hofstede’s Cultural Alignment Test \citep{hofstede2013vsm}, demonstrates that LLMs also struggle to represent cultural value systems beyond the dominant ones. Continuing in this direction, other works use survey instruments to define interpretable cultural spaces and to evaluate how LLM responses align with different national or cultural profiles \citep{durmus2023towards}. \citet
{tao2024cultural} used these surveys to formalize cultural distance, developing the Inglehart–Welzel Cultural Map derived from the World Values Survey (WVS) \citep{WVSCultMap} to locate LLMs culturally. However, assessing cultural alignment in LLMs remains inherently challenging. Measured alignment can vary substantially with prompt formulation \citep{khan2025randomness} and does not always correlate with human-perceived values \citep{arora2023probing}. Moreover, these approaches predominantly rely on question–answering or prompt-based evaluations, offering limited insight into how cultural knowledge is internally represented or operationalized during open-ended generation \citep{adilazuarda2024towards}. As a result, they primarily assess whether models can reproduce culturally conditioned responses rather than how cultural differences shape the creation of new content. Our work directly addresses this limitation by focusing on the generative production of cultural artifacts.

\subsection{Cultural Creativity, Novelty, and Distance}

Beyond values and beliefs, culture plays a central role in shaping how creativity and novelty are produced and perceived \citep{paletz2008implicit}. Creativity is commonly defined as the ability to create artifacts that are both novel and appropriate within a given cultural context \citep{boden2010creativity}. Cultural products such as music, literature, and cuisine are therefore key artifacts in which creativity manifests through the transformation of shared values and traditions rather than unconstrained invention \citep{appadurai1988make,guerrero2009consumer}. Moreover, perceptions of novelty and creativity are culturally situated, depending on both individual background and broader cultural context \citep{zhou2017new}. In particular, cultural distance has been shown to shape creative outcomes, as exposure to culturally distant ideas can increase perceived novelty while simultaneously challenging appropriateness or authenticity \citep{chua2015impact}. Creativity thus emerges at the intersection of novelty and cultural grounding, making cultural distance a critical factor in how creative transformation is evaluated. In computational linguistic research, cultural differences have similarly been studied as an emergent property of language use. Divergence between groups has been quantified using distributional measures such as relative entropy to capture differences in communication styles, linguistic norms, or value expression \citep{pechenick2015characterizing, klingenstein2014civilizing, gallagher2018divergent}. Building on this perspective, a recent work has proposed using textual novelty and divergence as proxies for cultural creativity in cooking recipes \citep{carichon2025crossing}. The authors operationalize cultural adaptation in human recipe adaptations, showing that linguistic divergence metrics correlate with established notions of cultural distance. Our work extends this framework to LLM-generated content, enabling direct comparison between human and model-generated cultural artifacts and allowing us to assess whether LLM productions reflect culturally grounded creativity.

\section{Methodology}

\subsection{Problem Statement}

In their article, \cite{carichon2025crossing} used the \textit{GlobalFusion} \footnote{\url{https://drive.google.com/file/d/1kYUw1BIym8E55gmloYLDFlkTJm71VxKi/view?usp=sharing}} dataset, composed of human-made recipes, to highlight the correlation between cultural distance and the notion of cultural novelty. The dataset consists of 500 dishes $D$, and for each dish $d_i \in D$, there are two sets of documents. A first set of reference recipes $Recipes$, composed of a recipe title, a set of ingredients, and set of $N$ textual instructions $T_i = {t_i^{1}; t_i^{2}, ..., t_i^{n}}$ acting as artifacts of a country's culture (e.g: Moroccan Couscous). Then a second set of $M$ variation or novel recipes, coming from different countries, $VC_i = {vc_{i}^{1}, vc_{i}^{2}, ..., vc_{i}^{m}}$, where $C$ represents the country of variation. While all dishes include, on average, variations from 19 countries, including those from the same country of origin (e.g., Moroccan Couscous, Jamaïcan Couscous, Mongolian Couscous, etc.), the total number of countries represented in the database is 130 from all continents.

Using this setup, for each dish we can measure the divergence between the reference recipes and the recipes from one variation country $C$, using information-theoretic metrics based on the Jensen-Shannon divergence, such as $Dist(R_i, VC_i)$. Then we can demonstrate the correlation between this textual divergence and notions of cultural distances, such as a Euclidean distance based on the Inglehart–Welzel Cultural Map \citep{WVSCultMap, tao2024cultural}, a linguistic distance\footnote{\url{http://dow.net.au/?page_id=32}}, a religious distance\footnote{\url{http://dow.net.au/?page_id=35}}, and a geographical distance between countries \citep{carichon2025crossing}.


\subsection{Dataset: LLMFusion}
\label{sec:dataset}

To extend \textit{GlobalFusion} such that we can compare cultural novelty between humans and LLMs, we first extract from each dish $d_i$ the set of countries $C_i$ associated with all the existing variations from that dish. For each dish,, we first have the standard prompt that requires generating a recipe from the dish's origin country $C^{orig}_i \in C_i$. Then, for each country of variation $C^{j}_i \in C_i$, we prompt an LLM to generate a novel recipe coming from that country. Figure \ref{prompt:instance_ex} presents one basic prompt example used to produce all the recipes. 

\begin{figure}
\begin{promptbox}
"Create a \textless KW\textgreater \textless NATIONALITY\textgreater version of this recipe: \textless RECIPE-NAME\textgreater. 

Please return, in English only, the following: 
1. A recipe title. 
2. A list of ingredients. 
3. A set of cooking instructions. 

The instructions must use only the ingredients listed above, be clear and concise, and maintain the structure and order described. Title:"
\end{promptbox}
\caption{Prompt example to generate multiple variations of a recipe given a specific NATIONALITY and a RECIPE-NAME.}
\label{prompt:instance_ex}
\end{figure}

\begin{figure}
    \centering 
    \includegraphics[width=0.9\linewidth]{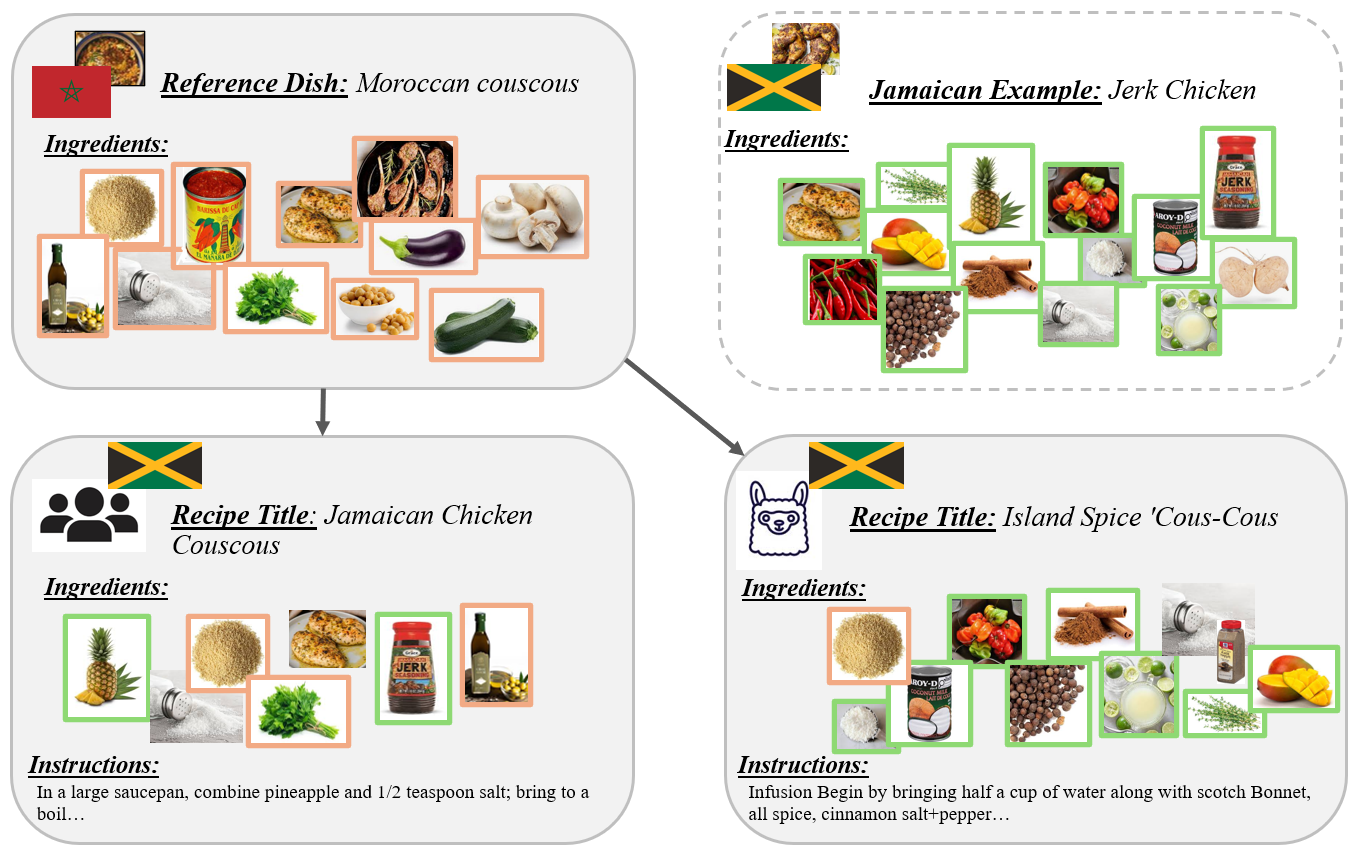}
    \caption{Example of a Jamaican Cultural Variations paired with Moroccan Couscous. The top left shows the ingredient set of Moroccan couscous reference recipes, while the top right displays ingredients characteristic of Jamaican cuisine, exemplified by Jerk Chicken. The bottom-right part presents human Jamaican couscous adaptations, and the bottom-left part presents an LLM-generated recipe using our protocol. While the human adaptation balances ingredients drawn from both Moroccan and Jamaican culinary traditions, the LLM-generated recipe predominantly relies on Jamaican ingredients, with minimal reuse of elements from the Moroccan reference.}
    \label{fig:recipes_example}
\end{figure}

\noindent Where \textless RECIPE-NAME\textgreater is the name of the dish $d_i$, and \textless NATIONALITY\textgreater can be equal to $\emptyset \; or  \; C^{orig}_i$ for the recipes coming from the same country of the dish, or equal to $C^{j}_i$ for each country of variation. <KW> is a set of keywords that forces the model to produce original, creative, or traditional recipes from different countries. The set \textless KW\textgreater is composed of the following words \{``novel'', ``unique'', ``new'', ``different'', ``surprising'', ``creative, desirable and useful'', ``original'', ``authentic'', ``traditional'', ``prototypical''\}. We also include, for each prompt, semantic variations to account for prompt sensitivity, adding explicit definitions and knowledge about keywords or the countries' cultural backgrounds. Additional prompt examples are provided in Appendix \ref{sec:app_prompts}. Finally, for each prompt, we ask the model to output recipes in English only and to follow the format: title, list of ingredients, and instructions, to be comparable with the human outputs in \textit{GlobalFusion}. We start the final prompts with the word "Title:" to guide the LLMs toward the desired output \citep{zhang2022automatic}. Therefore, at the end of this generative process, we obtain a total of 44 LLM-generated recipes for every country in $C_i$, and these recipes are all paired with existing variations created by humans.  Figure \ref{fig:recipes_example} provides an example of Jamaican variations from both human and LLMs paired with their origin recipes of Moroccan Couscous.

\subsection{Metrics}

To measure divergence in textual production, we rely on the metrics introduced in \citep{carichon2025crossing}. These metrics are based on the Jensen-Shannon divergence (JSD) and compare the information in a defined \textit{Culturally oriented knowledge space} with that in a new recipe called a cultural variation. Five different variations of the JSD are implemented to create five concepts of cultural divergence. Namely, the metrics are:
\begin{itemize}
\item Cultural Newness: The proportion of words significantly appearing or disappearing in the distribution of a text coming from a cultural variation compared to the distribution of the cultural knowledge base.
\item Cultural Uniqueness: The divergence measure of the distribution of a cultural variation compared to the prototypical view of the cultural product within a community. 
\item Cultural Difference: The distribution of a cultural variation is distant from all the observed points in the knowledge base associated with that cultural product. 
\item Cultural Surprise: The distribution of expected combinations of attributes of a cultural variation violates the projection of the cultural knowledge space distribution into a cultural expectation base. This notion declines in two sub-metrics: (1) the Cultural New Surprise, which relates to the sheer appearance of new combinations that did not exist in the expectation space, and (2) the Cultural Divergent Surprise, which measures the divergence between the PMI distribution of each term between our expectation space and our new observation.
\end{itemize}

We include additional details on the metrics' mathematical formulation in Appendix \ref{sec:app_metric}.

\subsection{Experiment}

To generate our results and recipes, we tested eight models across different sizes, training strategies, and multilingual capacities in zero-shot settings. The evaluated models include \textbf{Meta-Llama-3-70B-Instruct} and \textbf{gemma-2-27b-it}, which are English-only instruction-tuned models; \textbf{falcon-40b}, a non–instruction-tuned model trained on four languages; \textbf{Orion-14B-Chat}, a multilingual conversational model; \textbf{Phi-4-multimodal-instruct}, which supports both multilingual and multimodal inputs; \textbf{gemma-3-27b-it}, an instruction-tuned multimodal and multilingual model; \textbf{Qwen2.5-32B-Instruct}, a multilingual instruction-tuned model; and \textbf{Qwen3-30B-A3B-Instruct-2507}, a multilingual mixture-of-experts model with enhanced reasoning capabilities.

%

The text generation and analysis were performed using the Python library \textit{vLLMs}, and all models were downloaded from their HuggingFace version. The parameters used for the model are the following: we set the temperature for all models to 0 in order to have deterministic outputs. We use a penalty threshold of 1.4 to prevent token repetition in the generation output. The code to generate the recipe and score the novelty of each metric can be found in our GitHub repository\footnote{\url{https://github.com/fcarichon/LLMCokingNovelty}}.


\section{Results}

Before presenting our main results we highlight LLMs' ability to follow prompt instructions to generate recipes. Table \ref{tab:recipe_stats} introduces descriptive information about the generated recipes. While most instruction-based models easily follow the instruction, we observe performance degradation in the two other models, namely, orion-14b and falcon-40b, which also struggle to generate text long enough for good recipe instructions. However, the number of recipes remains sufficient to perform the remaining analysis. Overall, most recipes are generated in English, not too short (more than 50 tokens), and contain few repeated tokens (less than 3 repeated consecutive tokens), indicating low textual degradation. However, we also note that all LLMs struggle to use the ingredients listed in their instructions, and this tendency is even worse when asked to produce a cultural variation, where the ratio of properly used ingredients falls under 20\% for certain model.

\begin{table}[t]
\centering
\small
\setlength{\tabcolsep}{3pt}
\renewcommand{\arraystretch}{1.15}
\begin{tabular}{p{4.2cm}ccccccccc}
\hline
 & gemma-2 & gemma-3 & qwen-2 & qwen-3 & llama3-70b & phi4-8b & orion-14b & falcon-40b \\
\hline
Number of valid recipes 
& 197446 & 196116 & 189483 & 192731 & 196324 & 189062 & 22728 & 116887 \\

Average length of recipes 
& 114.21 & 274.028 & 392.63 & 425.24  & 199.066 & 303.05 & 65.6 & 60.64 \\

\% Instructions too short ($<50$ tokens) 
& 15.2 & 2.88 & 0.6 & 0.14 & 2.2 & 3.6 & 43.7 & 44.2 \\

\% of sentences with token repetition 
& 0.969 & 0.55 & 1.3 & 0.97 & 0.775 & 1.42 & 0.773 & 2.04 \\

\% of recipes in English 
& 99.3 & 95.7 & 98.42 & 95.27 & 98.6 & 99.03 & 98.1 & 99.2 \\

\hline
\end{tabular}
\caption{Descriptive characteristics of generated recipes across models.}
\label{tab:recipe_stats}
\end{table}

To investigate whether LLMs exhibit culturally grounded creativity, we structure our analyses around the following research questions. 
\begin{itemize}
    \item \textbf{RQ1.} Do LLM productions exhibit patterns of novelty that correlate with cultural distance, and how do these patterns compare to human adaptations?
    \item  \textbf{RQ2.} To what extent does the novelty in LLM recipes reflect creativity and traditions, rather than surface-level variation?
    \item \textbf{RQ3.} How do LLMs encode and ground cultural information, and how does it account for differences from human adaptations?
\end{itemize}
 
\subsection{RQ1: Cultural Distance in LLM and Human Recipes.}

In this section, we present our results on the relation between cultural novelty and cultural distance and we discuss the contrasts between human and LLM recipes. 

\subsubsection{Cultural novelty and cultural distance}
\begin{figure}
    \centering 
    \includegraphics[width=\linewidth]{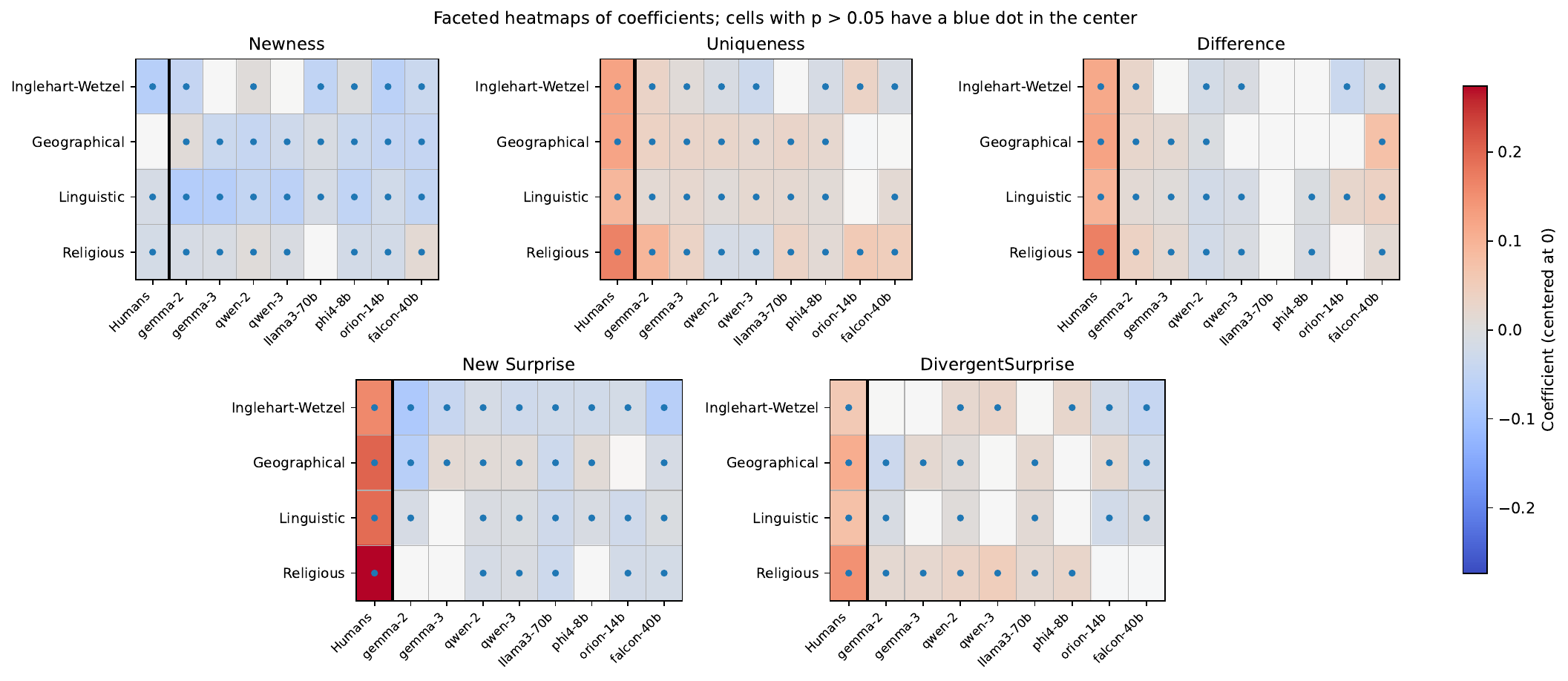}
    \caption{Correlations between the novelty metrics and the cultural distances for both human and LLMs. The color and its intensity provide information on the value of the coefficient, while the presence of the blue dot in each cell indicate a p value inferior to 0.05.}
    \label{fig:correl_analyses}
\end{figure}

\mbox{}\\
We study the Pearson correlations between cultural distances and divergence metrics. Figure \ref{fig:correl_analyses} presents our results.

We observe several systematic differences between human and model-generated recipes in the novelty metrics and their relationship to cultural distance. First, most models successfully reproduce, and even slightly reinforce, the correlation between \textit{Newness} and cultural distances comparable to those observed in human recipes. However, this metric also shows the weakest association with cultural distance and is not statistically significant in representing geographical distance in human recipes, indicating that \textit{Newness} alone is mainly independent of cultural variation. Second, models also capture a notion of \textit{Uniqueness}, but their correlation with cultural distance is substantially weaker than that of human recipes. A similar pattern is observed for the \textit{Divergent Surprise}, except that non-instruction-tuned models even exhibit an inverse correlation with cultural distance. Finally, for \textit{New Surprise}, all models invert the sign of the correlation with cultural distance relative to humans. The \textit{Difference} metric follows a similar pattern, except for models in the Gemma family; however, even in these cases, the strength of the association remains negligible compared to human recipes. Most importantly, \textit{New Surprise} and \textit{Difference} are precisely the two novelty metrics that exhibit the strongest correlations with cultural distance in human cooking recipes.

\subsubsection{Model Differences}
\mbox{}\\
While we observe quantitative differences across LLM families, these variations are mainly attributable to instruction-following capabilities rather than to multilingualism, multimodality, model scale, or more recent training regimes. In particular, Orion and Falcon exhibit markedly different behavior from other models: neither is instruction-tuned, and both generate substantially lower-quality recipes, with outputs averaging fewer than 60 tokens and frequently failing to follow the prompt constraints. As a result, their generations exhibit greater noise and inconsistency, making fine-grained cultural analysis more difficult and limiting the conclusions that can be drawn from their behavior.

Beyond these cases, the overall patterns remain consistent across models. We find no systematic difference between English-only and multilingual models, suggesting that exposure to multiple languages during training does not necessarily translate into improved cultural representativity in generative adaptation. However, as \textit{GlobalFusion} and our extended generative process are English-only recipes, this monolingual focus may limit multilingual models' capacity. Similarly, increased model capacity does not yield better alignment with human patterns: large models such as LLaMA-3-70B do not show stronger correlations with cultural distance than smaller counterparts. Finally, newer model versions optimized for safety, instruction-following, or reasoning, such as Gemma-3 and Qwen-3, do not exhibit meaningful improvements in capturing cultural divergence. Taken together, these observations indicate that current scaling strategies and training objectives largely overlook cultural diversity concerns. Despite architectural differences and successive model iterations, no LLM category consistently aligns more closely with human patterns of cultural adaptation.

 These findings show that although large language models produce recipes that appear novel in isolation, they fail to capture the culturally sensitive dimensions of novelty reflected in human adaptations.

\subsection{RQ2: LLM Creativity and Tradition vs. Surface Variations}

In this section, we provide an initial interpretation of why LLMs fail to align with cultural distance. In particular, we examine how LLMs produce cultural divergence and whether this reflects notions of creativity and traditionalism.

\subsubsection{LLMs overproduce divergence.}
\begin{figure}[t]
  \centering

  \begin{subfigure}{0.8\linewidth}
    \centering
    \includegraphics[width=\linewidth]{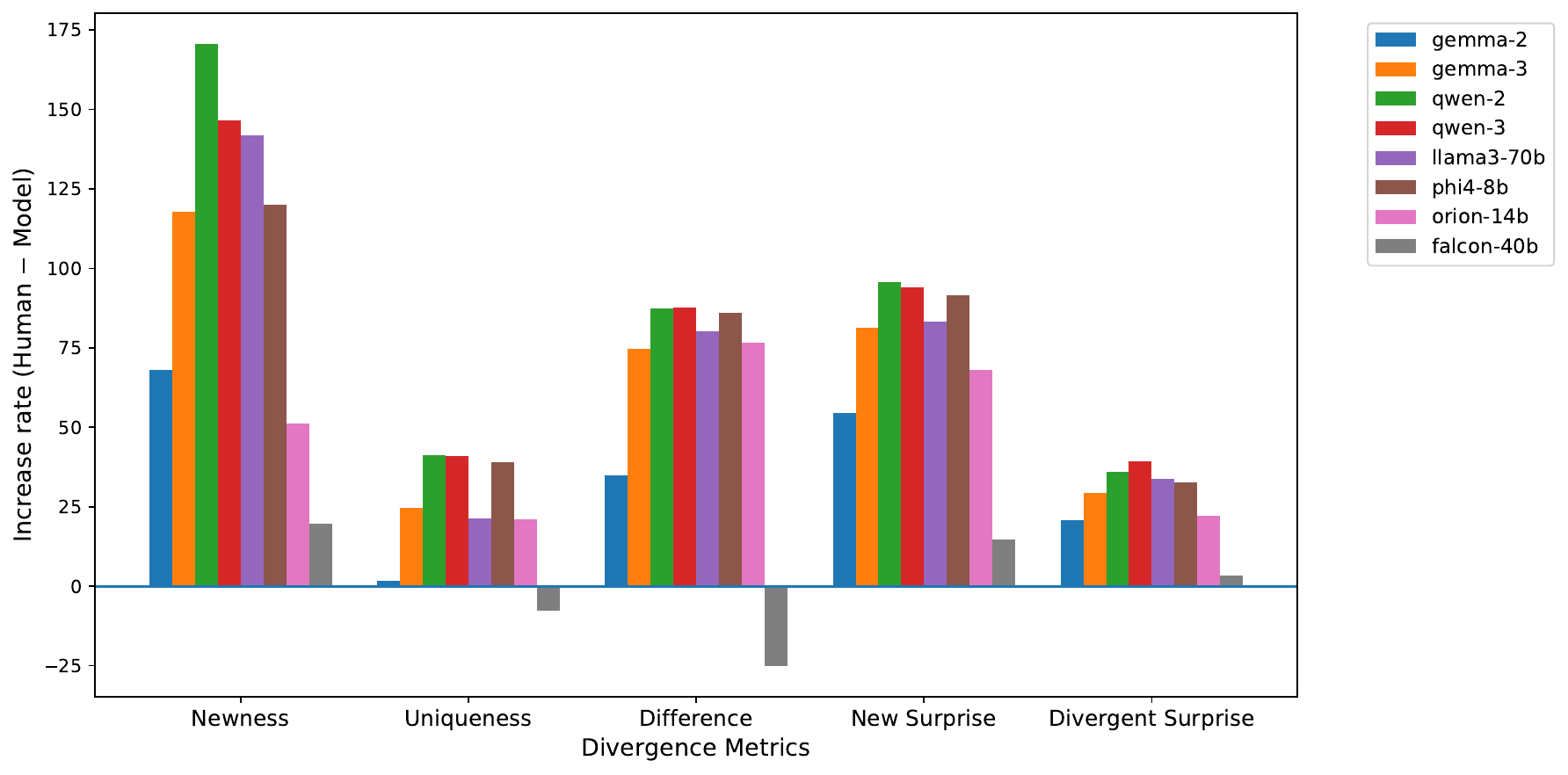}
    \caption{Country of Origin}
    \label{fig:hist_same}
  \end{subfigure}
  \hfill
  \begin{subfigure}{0.8\linewidth}
    \centering
    \includegraphics[width=\linewidth]{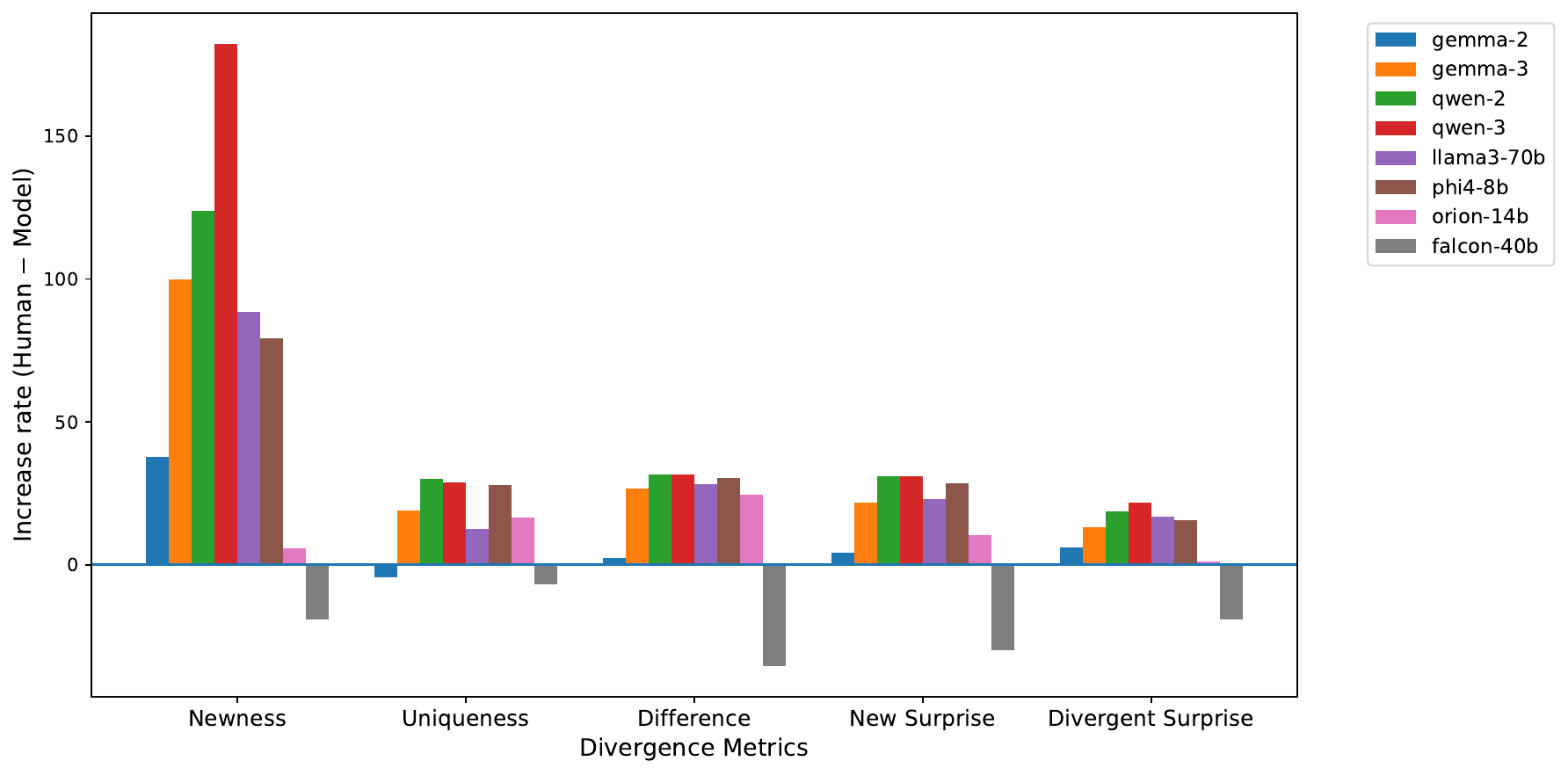}
    \caption{Paired Cultural Variations}
    \label{fig:hist_varia}
  \end{subfigure}

  \caption{Divergence increase rate between LLMs and human recipes. A positive value means that LLM-generated recipes have an increased divergence compared to human recipes.}
  \label{fig:hist_two_panels}
\end{figure}

\mbox{}\\
To better understand this dynamic, we are proposing several analyses comparing the divergence observed between human recipes and the LLM ones. First we compare the amount of divergence for each metric between human and models for recipes coming from the country of origin $C^{orig}_i$ of each dish and also for each pair of variation countries. Figure \ref{fig:hist_two_panels} presents the increase rate of the metrics for both cases.

We can observe that whether the recipe comes from the origin country or involves cultural variation, LLMs consistently overproduce divergence across all metrics and models. \textit{Newness} shows the highest increase rate compared to humans, and uniqueness and divergence are the closest between humans and models. Surprisingly, the only difference in behavior we observe comes from the Falcon-40b model, which shows less divergence than humans for some metrics. Further investigation would be needed to explain this difference for this specific model. Finally, we note that, for most metrics, the increase rate is lower across cultural variations, indicating that while human divergence increases, LLM divergence does not change. The only exception, once again, is \textit{Newness}, where the gap remains consistent, giving a first factor explaining why it is the only metric where LLMs preserve the same amount of correlation with cultural distance. 

\subsubsection{Limited Modeling of Creativity.}
\begin{figure}
    \centering 
    \includegraphics[width=0.7\linewidth]{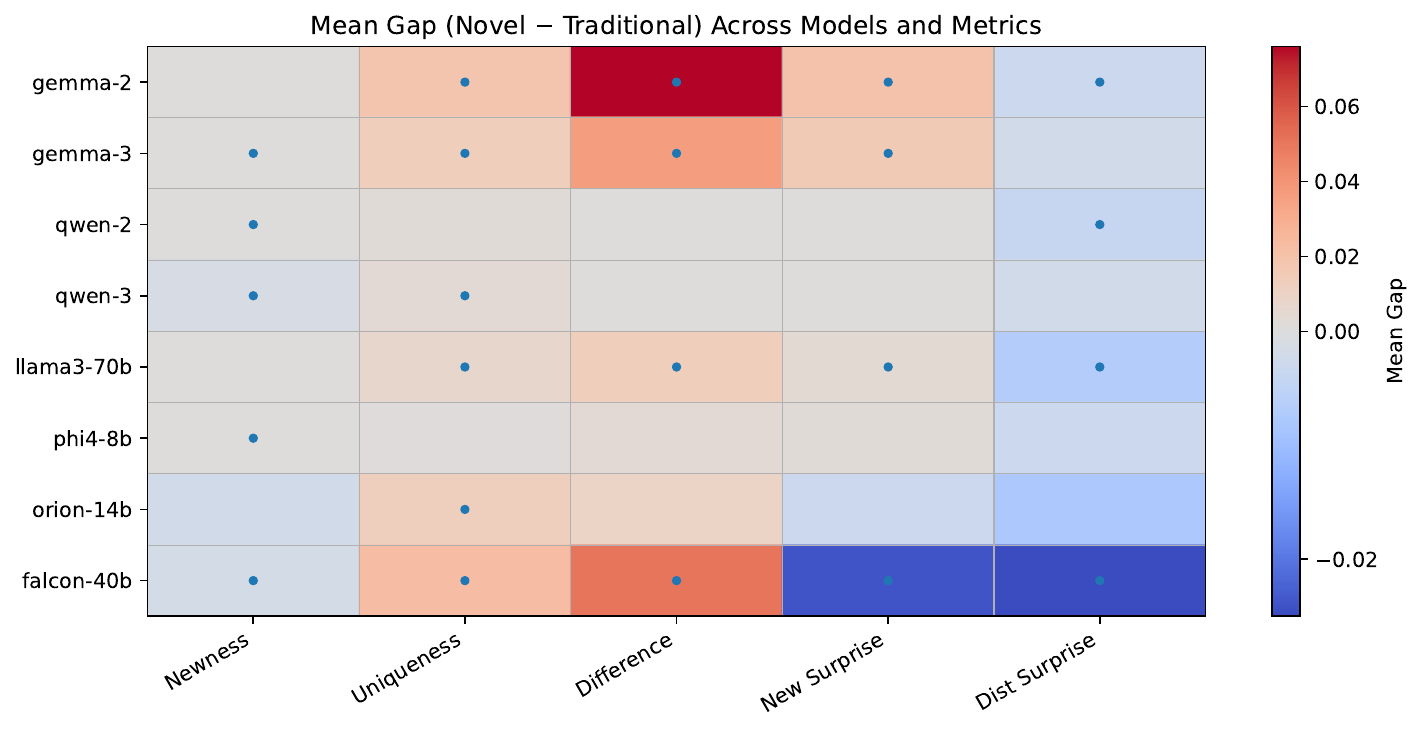}
    \caption{Average divergence gap between traditional recipes and novel and creative ones. The color and its intensity provide information on the value of the coefficient, while the presence of the blue dot in each cell indicate a p value inferior to 0.05.}
    \label{fig:tradi_novel}
\end{figure}

\mbox{}\\
Because cultural novelty and originality are closely tied to notions of authenticity and creativity, we further investigate how large language models respond to different keyword prompts related to these concepts. Specifically, we analyze the recipes generated by the models using the set of keywords introduced in Section \ref{sec:dataset}. We first focus on the recipes coming specifically from the country of origin $C^{orig}_i$ and we compare the amount of divergence when recipes were generated using traditional keywords such as ``authentic'', ``traditional'', ``prototypical'' versus the recipes generated with ''creative'' keywords such as ``novel'', ``unique'', ``new'', ``different'', ``surprising'', ``creative, desirable and useful''. Figure \ref{fig:tradi_novel} presents our results.

\begin{figure}
    \centering 
    \includegraphics[width=0.99\linewidth]{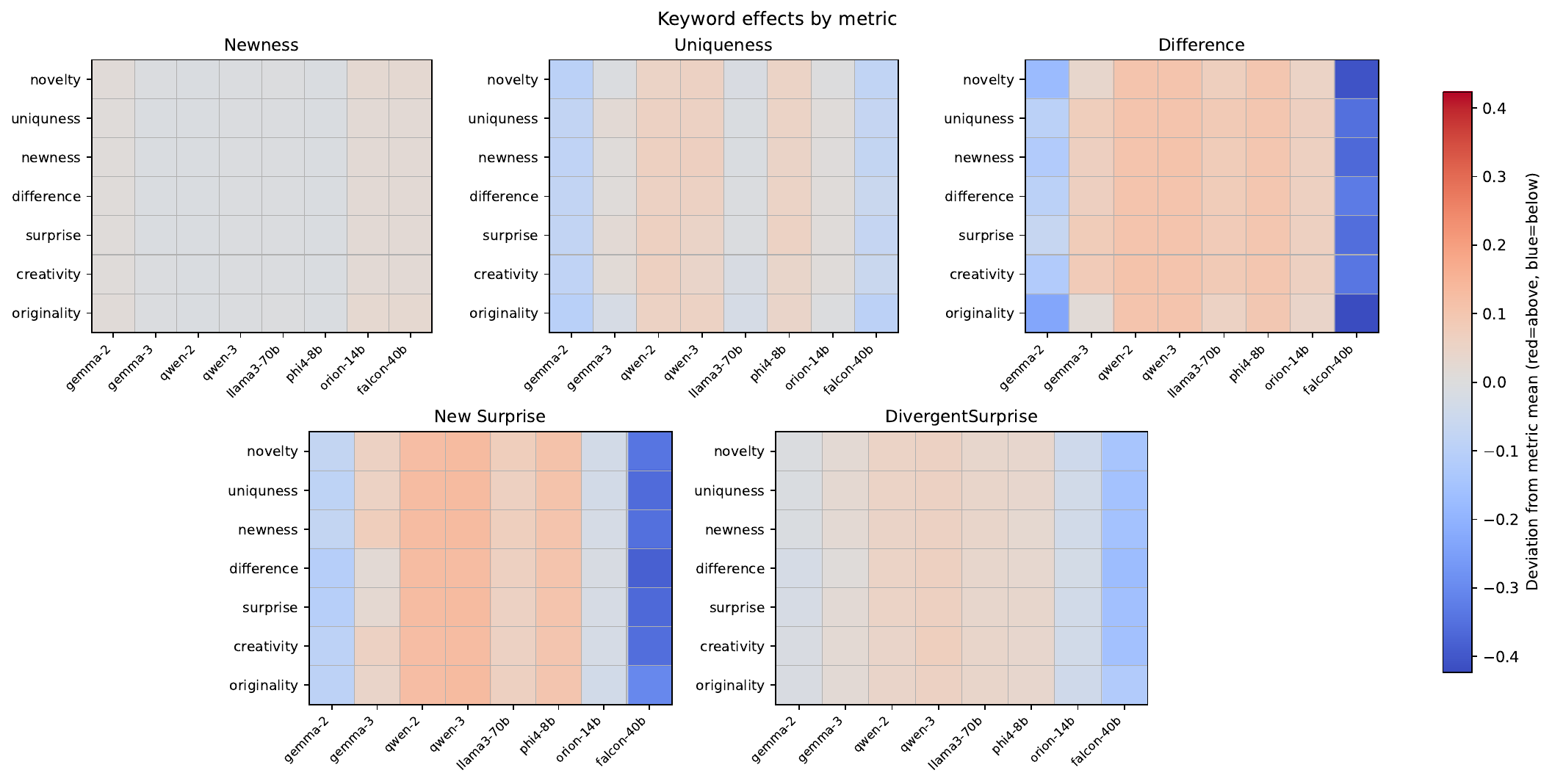}
    \caption{Divergences values for each metric with the set of ''creative'' keywords prompted when generating the recipes.}
    \label{fig:keywords_novel}
\end{figure}

The results show that prompting models with notions of novelty or creativity does not systematically lead to larger divergences than prompts associated with tradition, with effect sizes remaining small or model-dependent across the various metrics. Given the absence of a systematic increase in divergence between ''novelty'' and ''tradition'' oriented prompts, we further examine how models respond to different novelty-related keywords to characterize their sensitivity to these semantic distinctions. Therefore, we analyze model sensitivity to novelty-related prompting, for both variations and recipes from the original countries, by evaluating the effects of different keywords associated with novelty, creativity, and originality on divergence metrics. Figure \ref{fig:keywords_novel} presents our results. 

Across models, most metrics exhibit minimal variation across keywords, with consistently small effect sizes regardless of the semantic intent of the prompt. The metrics of \textit{Uniqueness} and \textit{Divergent Surprise} display moderate variation across models but remains weakly differentiated across keyword conditions. In contrast, \textit{Difference} and \textit{New Surprise}, reach high absolute values with several difference depending on models, but once again these metrics show limited sensitivity to the specific keyword used. In general, keywords seem to have no impact on textual divergence. With one exception, even if the difference are not statistically significant, we notice from the detailed table \ref{tab:kw_metrics} in Appendix \ref{sec:app_results} that the keywords of ''novelty'' and ''originality'' consistently produce the highest amount of \textit{Newness} and the lowest amount of \textit{Uniqueness} and \textit{Difference} across all models indicating a first pattern a the type of novelty induced by LLMs. Overall, variation across model architectures substantially exceeds variation across keyword prompts, indicating that novelty-related behavior is primarily driven by model-specific factors rather than semantic control through prompting.

As Figure \ref{fig:hist_two_panels} indicates, LLMs consistently outperform humans on novelty scores, especially on newness, both within the same country contexts and when adapting recipes across countries. This analysis, as well as the detailed keyword analysis provided in table \ref{tab:kw_metrics} in appendix \ref{sec:app_results}, suggests that for LLMs, novelty, particularly when explicitly prompted as ''originality'' or ''novelty'', is operationalized through the injection of new keywords and atypical semantic features that diverge from the core exemplar query \citep{peeperkorn2024temperature}. However, this mechanistic pursuit of novelty comes at a cost to cultural preservation. Because cultural specificity is often encoded in subtle, context-sensitive markers weakly correlated with the shear appearance of keywords, it is frequently de-prioritized or erased by the model's drive for "newness". Furthermore, we hypothesize that this over-valuation of ''newness'' artificially inflates ''surprise'' and ''difference'' metrics, decoupling them from human baselines. Consequently, the reliance on atypical (or low-frequency) terms likely breaks the correlation between novelty metrics and cultural distance, suggesting that LLMs achieve novelty by sacrificing cultural specificity.

\subsection{RQ3: LLMs Encoding and Grounding of Culture.}

This section offers a complementary perspective on the misalignment between LLM outputs and cultural distance by examining how cultural information is encoded and grounded in regional and ingredient factors.

\subsubsection{The loss of cultural encoding in LLM internal layers.}
\label{sec:layers}
\mbox{}\\

\begin{figure}
    \centering 
    \includegraphics[width=0.99\linewidth]{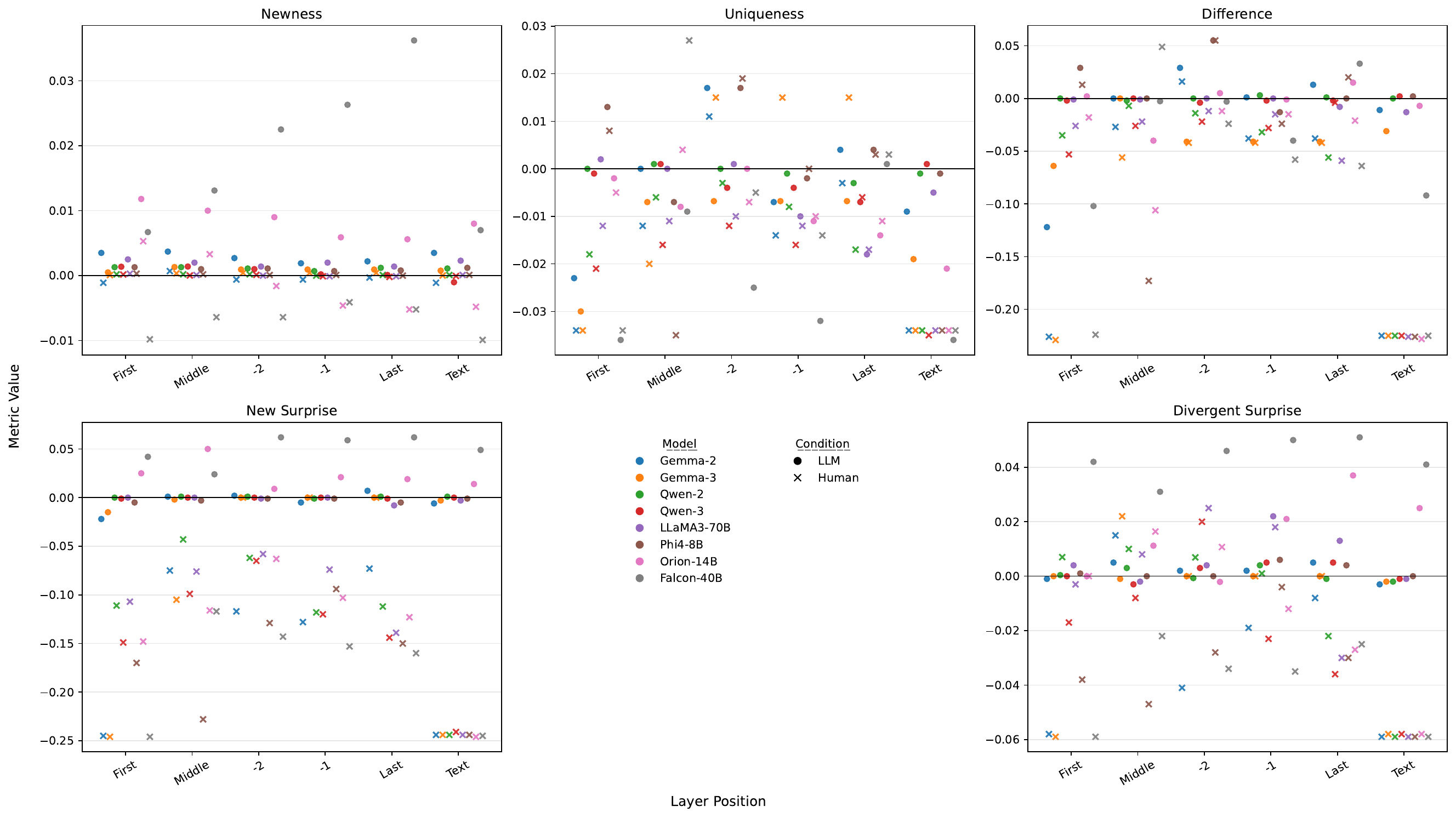}
    \caption{Evolution of the divergence average gap for each metric in several internal LLM layers. The gap considered indicates differences between recipes from the country of origin and those from cultural variations. Plain lines represent the levels for model-generated recipes, and dotted lines represent human recipes.}
    \label{fig:layer_div}
\end{figure}

A potential reason for the over-divergence and the lack of correlation with cultural distances in LLM outputs could be their ways of encoding this information. Therefore, we analyzed recipe divergence between humans and LLMs across the different layers of these models. More specifically, we follow the method introduced in \citep{li2025attributing} to compare the effect of RAG on internal layer representations. We first re-encode references, human, and model-generated recipes using the same LLM used for the initial generation. We then apply the \textit{Logit Lens} method \citep{nostalgebraist2020logitlens} to project intermediate hidden-layer representations into the token space. We can then compute divergence metrics over the decoded token distributions at each layer, enabling a direct comparison of how models internally represent human-authored versus self-generated recipes. Due to the computational cost of layer-wise analysis, we focus on five representative layers: the embedding layer, the middle layer, and the final three layers prior to text generation. Figure~\ref{fig:layer_div} summarizes the results across models and metrics. Either for human or for the model generated recipes, we presents the difference in divergence values when measured for recipes coming from the country of origin of the dish $C^{orig}_i$ with the values of divergences for recipes coming from cultural variations.

For newness, we observe a relatively stable pattern across all analyzed layers. Values remain small and vary only marginally from early to late layers, with similar trends observed across models. This stability is consistent across both LLM-generated and human recipes, indicating that newness-related divergence does not depend strongly on layer depth. 
In contrast, metrics that capture culturally conditioned divergence, namely \textit{Difference}, \textit{New surprise}, and \textit{Divergent Surprise}, exhibit a markedly different behavior. For these metrics, we observe information compression in the early and middle layers, suggesting limited separation between recipes from the origin country and their cultural variants at these stages. When examining representations derived from human recipes, higher divergence is observed, particularly in the later layers of the models. For LLM-generated recipes, the magnitude of this divergence remains substantially lower than that observed in human representations and varies across model families. Finally, the layer-wise behavior of \textit{Uniqueness} differs from the other measures. Across models, uniqueness values exhibit substantial variability across layers and architectures, with frequent sign changes and no consistent trend from early to late layers. It does not display a systematic separation between origin-country recipes and their cultural variations at any specific depth in the network. While some models show localized deviations, these effects are neither stable across layers nor consistent across model families. Overall, these results indicate that, for at least three of four culturally sensitive divergence metrics, we observe reduced divergence effects, suggesting that cultural differences are lost in the intermediate representations of text in LLMs, and that this loss of information is only partially recovered when producing recipes with such models.

These results demonstrate cultural representativity is not linearly preserved across the internal representation layers of LLMs. Instead, it is weakly encoded in early and middle layers and insufficiently reconstructed to support autonomous, divergent generation by LLMs.  These results are consistent with those observed in studies of multilingual LLMs, which have shown that LLMs rely solely on a few pivot languages, such as English, in the middle layers \citep{alabi2024hidden} before reconstructing outputs of underrepresented languages in the very last layers of the model \citep{bandarkar2025multilingual}. Our observations are also consistent with studies showing that models are losing information to balance compression and preservation of information for task performance within intermediate layers \citep{skean2025layer}, potentially leading to a flattening of cultural information towards the most dominant culture \citep{yu2025entangled}. We observe the same phenomenon here: cultural divergence is lost in the middle layers, and since the LLMs' compensation in the last layers is mainly superficial to ensure task completion (i.e., reproduce the human output), it is not sufficient to reproduce this type of divergence later. 

\subsubsection{The Problem of Cultural Tradition \& Authenticity}
\label{sec:trad_countries}
\begin{figure}[t]

  \begin{subfigure}{0.8\linewidth}
  \centering
  \includegraphics[width=0.95\linewidth]{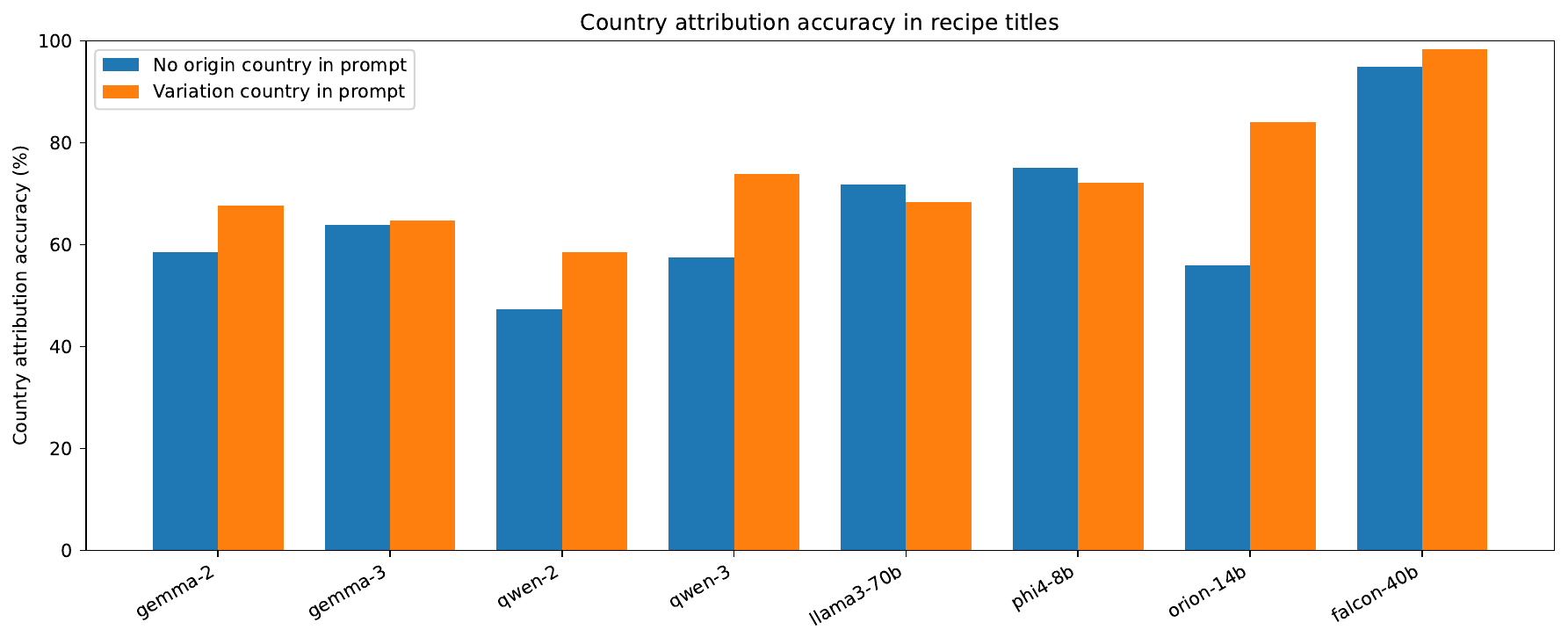}
  \caption{Proportion of country mismatches}
  \end{subfigure}
  \hfill
  \begin{subfigure}{0.8\linewidth}
  \centering
  \includegraphics[width=0.7\linewidth]{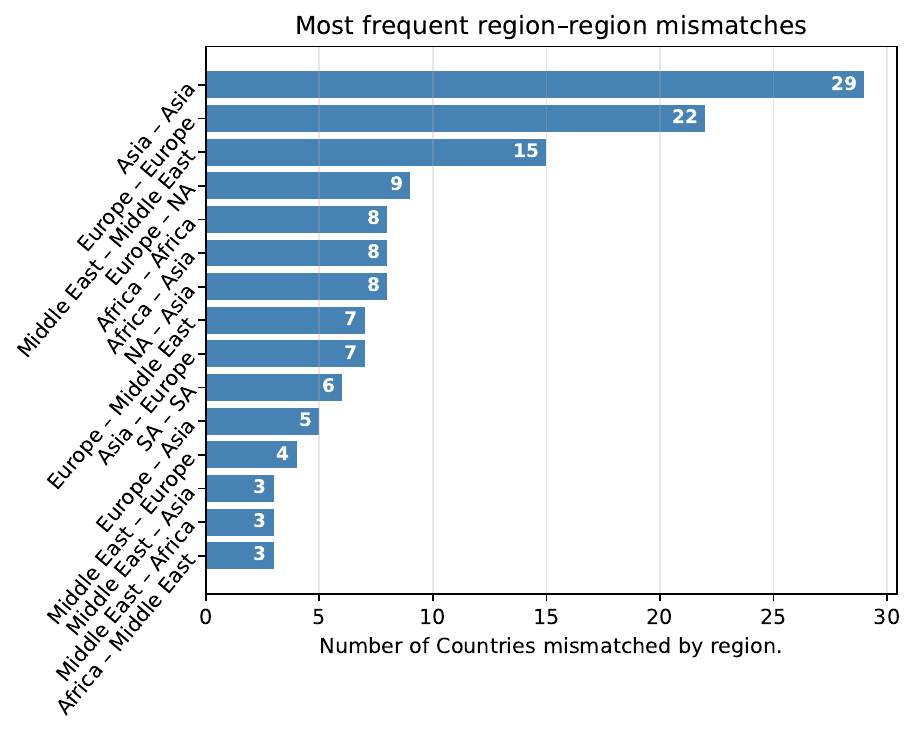}
  \caption{Pairs of region being mismatch at least one by models}
  \end{subfigure}

  \caption{Country attribution accuracy in recipe title. The graphs highlights the propensity of errors made by model in recipe titles.}
  \label{fig:country_mismatch}
\end{figure}

\mbox{}\\
Food and cuisine encode cultural traditions that both represent group identity and delineate boundaries between cultures \citep{appadurai1988make, hobsbawm2012invention}. However, as we observed in Figure \ref{fig:tradi_novel}, models are almost unable to produce divergence when dealing with novel recipes or traditional or authentic ones. We think one of the main reasons is their misunderstanding of these boundaries. To provide concrete evidence for this hypothesis, we analyzed the recipe titles and specifically whether LLMs mismatched country attributions for specific recipes. More specifically, we looked at prompts where we did not provide any information about the dish's country of origin and at which country was mentioned in the recipe title. We also examined the explicit mismatch in cultural variation when the country of variation is mentioned in the prompt. Figure \ref{fig:country_mismatch} presents our results. 

We observe that country attribution is frequently incorrect, with 25 to 50\% of errors when the country is not explicitly mentioned and 15 to 40\% when the country is specified. These errors are not random, as they cluster around popular countries and cuisines. The most frequent countries in these mismatches are South Korea, Morocco, Greece, Thailand, Italy, France, China, and the US, the most represented countries in \textit{GlobalFusion}. Moreover, most mismatches occur within the same region (e.g., Asia–Asia, Europe–Europe), indicating blurred cultural boundaries. For example, the most common mistake we see in the dataset is Taiwan and Japan being replaced by China, Spain by Italy or Mexico, and Tunisia by Morocco. These observations tend to confirm that LLMs novelty is not anchored to cultural identity, and therefore cannot produce divergence between traditional and creative recipes that would be culturally meaningful.

\subsubsection{Difference in Material Grounding of Cultural Novelty}
\label{sec:mat_ground}
\mbox{}\\
Finally, this suggests that LLM outputs are not totally grounded in the cultural artifacts that make recipes representative of specific cultures or countries. Prior work shows that ingredient modification play a central role in culinary adaptation \citep{guerrero2009consumer}. Therefore we also examine how LLMs use ingredients in generated recipes to underlie this discrepancy.

\begin{figure}
    \centering 
    \includegraphics[width=0.85\linewidth]{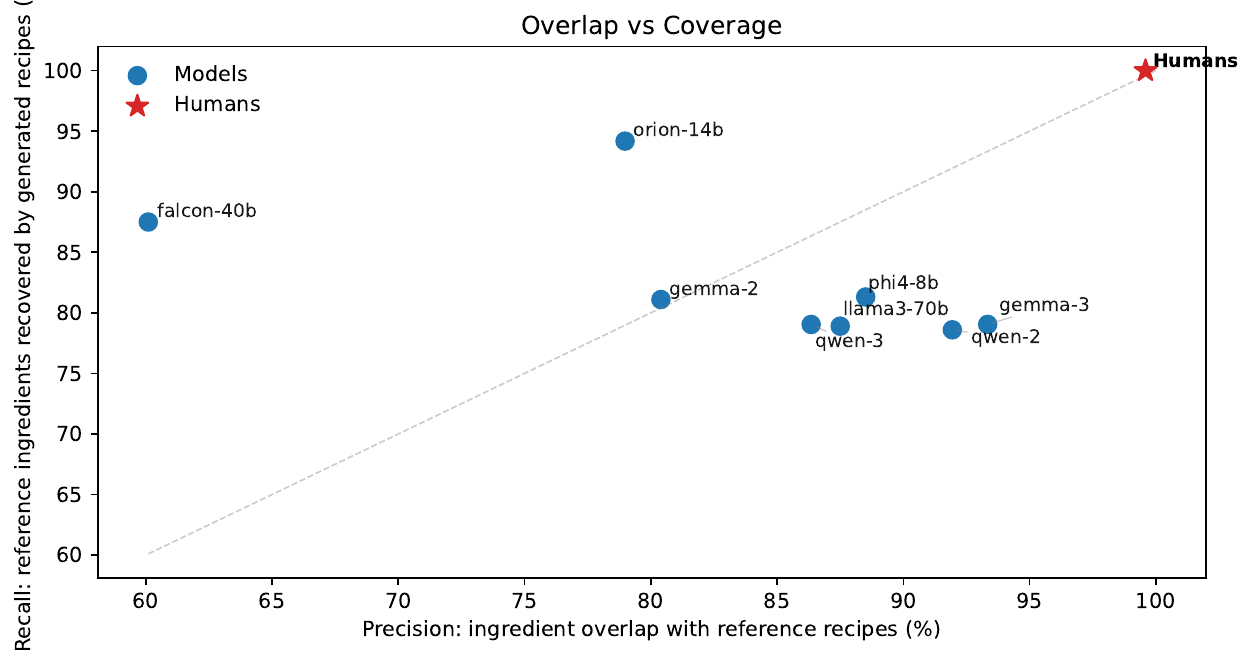}
    \caption{Ingredient overlap and preservation between human and LLM-generated recipes.}
    \label{fig:ingr_overlap}
\end{figure}

We first looked at the recipes from the country of origin of the dish. We first measure precision as the \textbf{overlap} between the ingredients in recipes, either for LLMs or humans, and the list of all ingredients used in human references for this dish. We also examined a notion of recall as the proportion of reference ingredients from the country of origin \textbf{preserved} in LLM- generated or human recipes. Figure \ref{fig:ingr_overlap} presents our results. What we first notice is that humans preserve almost all ingredients and adapt recipes without discarding the cultural core, confirming that novelty is made through means other than just ingredient replacement \citep{guerrero2009consumer}. In contrast, we observed a greater overlap and lesser coverage of the core ingredients in LLM recipes. This asymmetry suggests that LLMs mostly use ingredients that appear in human references, but they systematically fail to recover many ingredients that humans consider essential to a dish. Falcon-40B and Orion-14B constitute notable exceptions, exhibiting inverse overlap patterns that are consistent with their tendency to generate noisy or malformed recipes. Moreover, we provide regional details on the overlap and coverage of ingredients in Table \ref{tab:regional_ingr_overlap} in Appendix \ref{sec:app_results}, which highlights another consistent pattern across models. East Asia and South America have the lowest coverage; meanwhile, regions such as Europe, North America, and Asia perform better. Therefore, regions that are less represented or more culturally distinctive in ingredients suffer the largest loss of components. It seems the reason is that LLMs, when producing novel recipes, expand without grounding and default to generic, or stereotypical ingredients.

The analysis in Table \ref{tab:top20_ingredients} in Appendix \ref{sec:app_results} presenting the most frequently used ingredients in LLM-generated recipes further supports this interpretation. We notice that the generation converges toward a small set of globally ubiquitous ingredients such as onion, garlic, salt, pepper, oil, sugar, and flour, largely independent of any culture. Moreover, LLMs substitute culturally grounded specificities (e.g. black pepper, Sichuan pepper, chili pepper or semolina, rice flour, cassava flour) with procedural placeholders (e.g., “salt to taste”, “optional”), indicating this shift toward generic recipes.

\begin{figure}
    \centering 
    \includegraphics[width=0.99\linewidth]{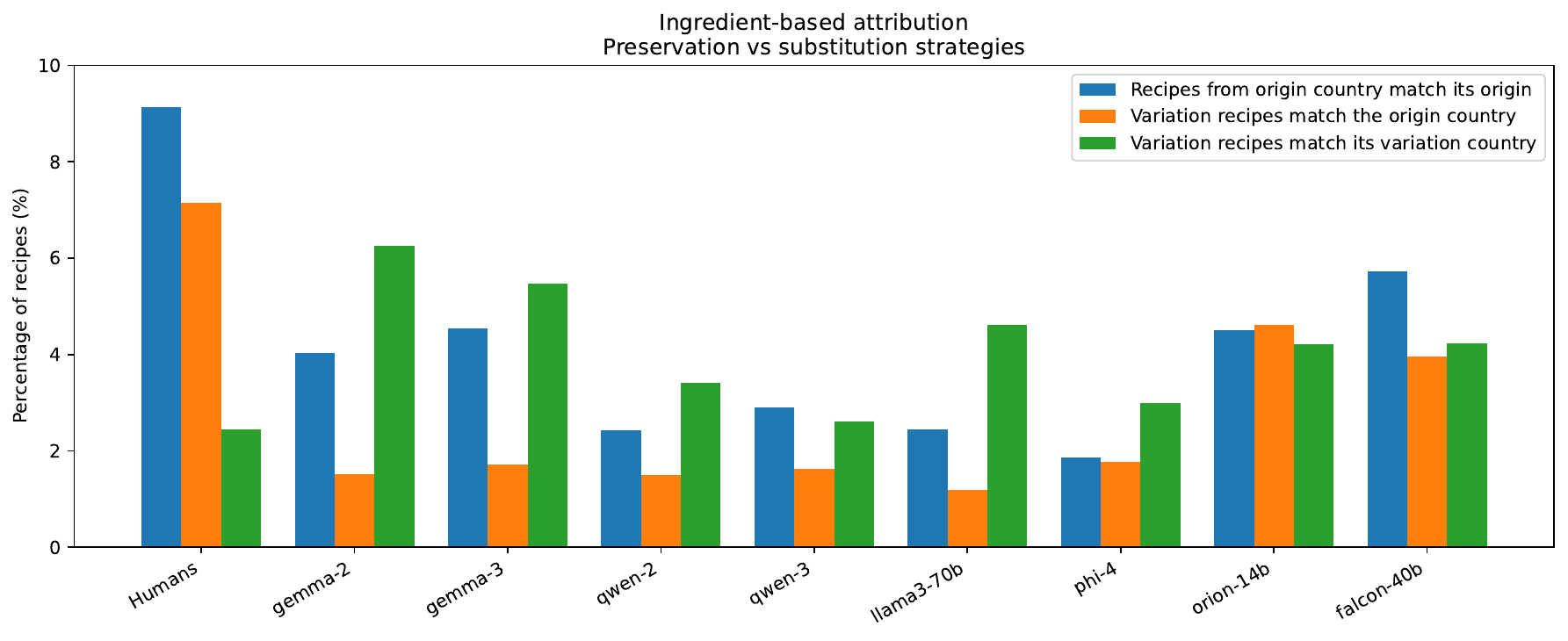}
    \caption{Ingredient-based cultural attribution using TF-IDF cosine similarity. We report the proportion of recipes whose ingredient profiles most closely match their country of origin or their associated cultural variation.}
    \label{fig:ingr_attribution}
\end{figure}

Finally, even when ingredients overlap with their associated country, the question remains on the representativeness of those ingredients to the country's culinary identity. To address this, we analyze ingredient typicality by modeling country-specific ingredient distributions using a TF-IDF representation. Based on ingredients TFIDF scores, we then create representations for the different countries and new human or LLM-generated recipes. Finally, we measure the cosine similarity between these recipes and the countries to quantify whether generated recipes are more strongly associated with their country of origin, their target variation, or neither. Figure \ref{fig:ingr_attribution} presents our results. We observe that human recipes align most closely with their country of origin, both for original recipes and for culturally adapted variants, indicating once again that adaptation typically preserves culturally core ingredients. In contrast, LLM-generated recipes exhibit substantially weaker alignment with origin countries and shift toward target variation cuisines reflecting once again the propensity of LLMs to substitute rather than preserve ingredients. The example in Figure \ref{fig:recipes_example} displaying a Jamaican variation of the Moroccan Couscous highlights this tendency. While the LLM recipe include a lot of ingredients coming from Jamaican Cuisine, the human one is more balanced on the ingredients used. 

\section{Limitations}

Our work has several limitations for its analyses and interpretations. First, our analysis measures divergence in LLM-generated recipes relative to human references rather than LLM reference baselines. While this choice is motivated by our focus on human cultural adaptation, it implies that observed divergences are evaluated against human norms of culinary variation and may not fully reflect how models internally structure cultural knowledge. In addition, our study relies on the implicit cultural representations learned by LLMs during pretraining rather than on explicit, curated knowledge bases, which may introduce biases related to data availability and training distributions.

Second, our results inherit the limitations from the \textit{GlobalFusion} dataset, including an uneven representation across countries and regions. Although the dataset spans 130 countries, it disproportionately reflects cuisines that are frequently documented online, which may amplify observed disparities for less represented or culturally specific regions.

Finally, our analysis is restricted to English-only recipes. This hinders cultural representativity in both human cooking recipes, as these recipes are often translated online and LLMs since it influences LLMs alignment with cultural values \citep{cao2023assessing}. Moreover, several of the evaluated models are multilingual, and we do not explicitly assess how language choice interacts with specific cultural grounding. Incorporating multilingual recipe data could reveal different adaptation strategies and remains an important direction for future work. 

\section{Conclusion}

Our results show that current LLMs do not reproduce culturally grounded creative patterns observed in human practices, generating novelty that is weakly correlated with cultural distance and insufficiently anchored in tradition or culturally salient elements. By analyzing divergence, internal representations, and ingredient usage, we identify structural limitations that contribute to the lack of cultural representation in generative output. These findings underscore the need for artifact-based evaluation frameworks and more culturally aware model designs when deploying LLMs in culturally sensitive contexts.

\section*{Acknowledgments}
Funding support for project activities has been partially provided by the Canada CIFAR AI Chair, an IVADO award and an Open Philanthropy award. This research was enabled in part by computing resources provided by Mila and the Digital Research Alliance of Canada.

\section*{Generative AI Usage Statement}
In the preparation of this article, we used AI-assisted writing tools (ChatGPT v5.2, Grammarly) to support sentence rephrasing, table formatting, and grammatical refinement.

\bibliographystyle{apalike}  
\bibliography{references}

@String{Springer = "Springer-Verlag" }

@article{atari2023humans,
  title={Which humans?},
  author={Atari, Mohammad and Xue, Mona and Park, Peter and Blasi, Dami{\'a}n and Henrich, Joseph},
  year={2023},
  publisher={OSF}
}

@inproceedings{carichon2025crossing,
  title={Crossing Boundaries: Leveraging Semantic Divergences to Explore Cultural Novelty in Cooking Recipes},
  author={Carichon, Florian and Rampa, Romain and Farnadi, Golnoosh},
  booktitle={Proceedings of the 2025 ACM Conference on Fairness, Accountability, and Transparency},
  pages={2200--2214},
  year={2025}
}

@misc{WVSCultMap,
  title = {The Inglehart-Welzel World Cultural Map - World Values Survey 7 (2023).},
  howpublished = {\url{https://www.worldvaluessurvey.org/WVSContents.jsp}},
  author = {World Values Survey (WVS)},
  note = {Dataset collection date:: 2017-2022}
}

@article{tao2024cultural,
  title={Cultural bias and cultural alignment of large language models},
  author={Tao, Yan and Viberg, Olga and Baker, Ryan S and Kizilcec, Ren{\'e} F},
  journal={PNAS nexus},
  volume={3},
  number={9},
  pages={pgae346},
  year={2024},
  publisher={Oxford University Press US}
}

@article{zhang2022automatic,
  title={Automatic chain of thought prompting in large language models},
  author={Zhang, Zhuosheng and Zhang, Aston and Li, Mu and Smola, Alex},
  journal={arXiv preprint arXiv:2210.03493},
  year={2022}
}

@article{li2025attributing,
  title={Attributing Response to Context: A Jensen-Shannon Divergence Driven Mechanistic Study of Context Attribution in Retrieval-Augmented Generation},
  author={Li, Ruizhe and Chen, Chen and Hu, Yuchen and Gao, Yanjun and Wang, Xi and Yilmaz, Emine},
  journal={arXiv preprint arXiv:2505.16415},
  year={2025}
}

@misc{nostalgebraist2020logitlens,
  author       = {nostalgebraist},
  title        = {Interpreting GPT: The Logit Lens},
  year         = {2020},
  month        = aug,
  howpublished = {\url{https://www.lesswrong.com/posts/AcKRB8wDpdaN6v6ru/interpreting-gpt-the-logit-lens}},
  note         = {Accessed August 2020}
}

@article{guerrero2009consumer,
  title={Consumer-driven definition of traditional food products and innovation in traditional foods. A qualitative cross-cultural study},
  author={Guerrero, Luis and Gu{\`a}rdia, Maria Dolors and Xicola, Joan and Verbeke, Wim and Vanhonacker, Filiep and Zakowska-Biemans, Sylwia and Sajdakowska, Marta and Sulmont-Ross{\'e}, Claire and Issanchou, Sylvie and Contel, Michele and others},
  journal={Appetite},
  volume={52},
  number={2},
  pages={345--354},
  year={2009},
  publisher={Elsevier}
}

@article{skean2025layer,
  title={Layer by layer: Uncovering hidden representations in language models},
  author={Skean, Oscar and Arefin, Md Rifat and Zhao, Dan and Patel, Niket and Naghiyev, Jalal and LeCun, Yann and Shwartz-Ziv, Ravid},
  journal={arXiv preprint arXiv:2502.02013},
  year={2025}
}

@article{yu2025entangled,
  title={Entangled in representations: Mechanistic investigation of cultural biases in large language models},
  author={Yu, Haeun and Jeong, Seogyeong and Pawar, Siddhesh and Shin, Jisu and Jin, Jiho and Myung, Junho and Oh, Alice and Augenstein, Isabelle},
  journal={arXiv preprint arXiv:2508.08879},
  year={2025}
}

@article{peeperkorn2024temperature,
  title={Is temperature the creativity parameter of large language models?},
  author={Peeperkorn, Max and Kouwenhoven, Tom and Brown, Dan and Jordanous, Anna},
  journal={arXiv preprint arXiv:2405.00492},
  year={2024}
}

@article{appadurai1988make,
  title={How to make a national cuisine: cookbooks in contemporary India},
  author={Appadurai, Arjun},
  journal={Comparative studies in society and history},
  volume={30},
  number={1},
  pages={3--24},
  year={1988},
  publisher={Cambridge University Press}
}

@book{hobsbawm2012invention,
  title={The invention of tradition},
  author={Hobsbawm, Eric and Ranger, Terence},
  year={2012},
  publisher={Cambridge university press}
}

@article{pawar2025survey,
  title={Survey of cultural awareness in language models: Text and beyond},
  author={Pawar, Siddhesh and Park, Junyeong and Jin, Jiho and Arora, Arnav and Myung, Junho and Yadav, Srishti and Haznitrama, Faiz Ghifari and Song, Inhwa and Oh, Alice and Augenstein, Isabelle},
  journal={Computational Linguistics},
  pages={1--96},
  year={2025},
  publisher={MIT Press 255 Main Street, 9th Floor, Cambridge, Massachusetts 02142, USA~…}
}

@inproceedings{masoud2025cultural,
  title={Cultural alignment in large language models: An explanatory analysis based on hofstede’s cultural dimensions},
  author={Masoud, Reem and Liu, Ziquan and Ferianc, Martin and Treleaven, Philip C and Rodrigues, Miguel Rodrigues},
  booktitle={Proceedings of the 31st International Conference on Computational Linguistics},
  pages={8474--8503},
  year={2025}
}

@inproceedings{khan2025randomness,
  title={Randomness, not representation: The unreliability of evaluating cultural alignment in llms},
  author={Khan, Ariba and Casper, Stephen and Hadfield-Menell, Dylan},
  booktitle={Proceedings of the 2025 ACM Conference on Fairness, Accountability, and Transparency},
  pages={2151--2165},
  year={2025}
}

@article{cao2023assessing,
  title={Assessing cross-cultural alignment between ChatGPT and human societies: An empirical study},
  author={Cao, Yong and Zhou, Li and Lee, Seolhwa and Cabello, Laura and Chen, Min and Hershcovich, Daniel},
  journal={arXiv preprint arXiv:2303.17466},
  year={2023}
}

@article{johnson2022ghost,
  title={The Ghost in the Machine has an American accent: value conflict in GPT-3},
  author={Johnson, Rebecca L and Pistilli, Giada and Men{\'e}dez-Gonz{\'a}lez, Natalia and Duran, Leslye Denisse Dias and Panai, Enrico and Kalpokiene, Julija and Bertulfo, Donald Jay},
  journal={arXiv preprint arXiv:2203.07785},
  year={2022}
}

@article{mazeika2025utility,
  title={Utility engineering: Analyzing and controlling emergent value systems in ais},
  author={Mazeika, Mantas and Yin, Xuwang and Tamirisa, Rishub and Lim, Jaehyuk and Lee, Bruce W and Ren, Richard and Phan, Long and Mu, Norman and Khoja, Adam and Zhang, Oliver and others},
  journal={arXiv preprint arXiv:2502.08640},
  year={2025}
}

@article{adilazuarda2024towards,
  title={Towards measuring and modeling" culture" in llms: A survey},
  author={Adilazuarda, Muhammad Farid and Mukherjee, Sagnik and Lavania, Pradhyumna and Singh, Siddhant and Aji, Alham Fikri and O'Neill, Jacki and Modi, Ashutosh and Choudhury, Monojit},
  journal={arXiv preprint arXiv:2403.15412},
  year={2024}
}

@article{cao2024cultural,
  title={Cultural adaptation of recipes},
  author={Cao, Yong and Kementchedjhieva, Yova and Cui, Ruixiang and Karamolegkou, Antonia and Zhou, Li and Dare, Megan and Donatelli, Lucia and Hershcovich, Daniel},
  journal={Transactions of the Association for Computational Linguistics},
  volume={12},
  pages={80--99},
  year={2024},
  publisher={MIT Press One Broadway, 12th Floor, Cambridge, Massachusetts 02142, USA~…}
}

@article{hofstede2013vsm,
  title={VSM 2013},
  author={Hofstede, Geert and Minkov, Michael},
  journal={Values survey module},
  year={2013}
}

@article{durmus2023towards,
  title={Towards measuring the representation of subjective global opinions in language models},
  author={Durmus, Esin and Nguyen, Karina and Liao, Thomas I and Schiefer, Nicholas and Askell, Amanda and Bakhtin, Anton and Chen, Carol and Hatfield-Dodds, Zac and Hernandez, Danny and Joseph, Nicholas and others},
  journal={arXiv preprint arXiv:2306.16388},
  year={2023}
}

@article{hershcovich2022challenges,
  title={Challenges and strategies in cross-cultural NLP},
  author={Hershcovich, Daniel and Frank, Stella and Lent, Heather and de Lhoneux, Miryam and Abdou, Mostafa and Brandl, Stephanie and Bugliarello, Emanuele and Piqueras, Laura Cabello and Chalkidis, Ilias and Cui, Ruixiang and others},
  journal={arXiv preprint arXiv:2203.10020},
  year={2022}
}

@inproceedings{arora2023probing,
  title={Probing pre-trained language models for cross-cultural differences in values},
  author={Arora, Arnav and Kaffee, Lucie-aim{\'e}e and Augenstein, Isabelle},
  booktitle={Proceedings of the first workshop on cross-cultural considerations in NLP (C3NLP)},
  pages={114--130},
  year={2023}
}

@inproceedings{hu2024bridging,
  title={Bridging cultures in the kitchen: A framework and benchmark for cross-cultural recipe retrieval},
  author={Hu, Tianyi and Maistro, Maria and Hershcovich, Daniel},
  booktitle={Proceedings of the 2024 Conference on Empirical Methods in Natural Language Processing},
  pages={1068--1080},
  year={2024}
}

@article{paletz2008implicit,
  title={Implicit theories of creativity across cultures: Novelty and appropriateness in two product domains},
  author={Paletz, Susannah BF and Peng, Kaiping},
  journal={Journal of cross-cultural psychology},
  volume={39},
  number={3},
  pages={286--302},
  year={2008},
  publisher={Sage Publications Sage CA: Los Angeles, CA}
}

@article{shenkar2001cultural,
  title={Cultural distance revisited: Towards a more rigorous conceptualization and measurement of cultural differences},
  author={Shenkar, Oded},
  journal={Journal of international business studies},
  volume={32},
  pages={519--535},
  year={2001},
  publisher={Springer}
}

@article{pechenick2015characterizing,
  title={Characterizing the Google Books corpus: Strong limits to inferences of socio-cultural and linguistic evolution},
  author={Pechenick, Eitan Adam and Danforth, Christopher M and Dodds, Peter Sheridan},
  journal={PloS one},
  volume={10},
  number={10},
  pages={e0137041},
  year={2015},
  publisher={Public Library of Science San Francisco, CA USA}
}

@article{xie2023next,
  title={The next chapter: A study of large language models in storytelling},
  author={Xie, Zhuohan and Cohn, Trevor and Lau, Jey Han},
  journal={arXiv preprint arXiv:2301.09790},
  year={2023}
}

@article{liu2025wavjourney,
  title={Wavjourney: Compositional audio creation with large language models},
  author={Liu, Xubo and Zhu, Zhongkai and Liu, Haohe and Yuan, Yi and Huang, Qiushi and Cui, Meng and Liang, Jinhua and Cao, Yin and Kong, Qiuqiang and Plumbley, Mark D and others},
  journal={IEEE Transactions on Audio, Speech and Language Processing},
  year={2025},
  publisher={IEEE}
}

@article{meincke2024using,
  title={Using large language models for idea generation in innovation},
  author={Meincke, Lennart and Girotra, Karan and Nave, Gideon and Terwiesch, Christian and Ulrich, Karl T},
  journal={The Wharton School Research Paper Forthcoming},
  volume={9},
  pages={2024},
  year={2024},
  publisher={The Wharton School, University of Pennsylvania Philadelphia, PA, USA}
}

@article{brinkmann2023machine,
  title={Machine culture},
  author={Brinkmann, Levin and Baumann, Fabian and Bonnefon, Jean-Fran{\c{c}}ois and Derex, Maxime and M{\"u}ller, Thomas F and Nussberger, Anne-Marie and Czaplicka, Agnieszka and Acerbi, Alberto and Griffiths, Thomas L and Henrich, Joseph and others},
  journal={Nature Human Behaviour},
  volume={7},
  number={11},
  pages={1855--1868},
  year={2023},
  publisher={Nature Publishing Group UK London}
}

@article{gallagher2018divergent,
  title={Divergent discourse between protests and counter-protests:\# BlackLivesMatter and\# AllLivesMatter},
  author={Gallagher, Ryan J and Reagan, Andrew J and Danforth, Christopher M and Dodds, Peter Sheridan},
  journal={PloS one},
  volume={13},
  number={4},
  pages={e0195644},
  year={2018},
  publisher={Public Library of Science San Francisco, CA USA}
}

@article{klingenstein2014civilizing,
  title={The civilizing process in London’s Old Bailey},
  author={Klingenstein, Sara and Hitchcock, Tim and DeDeo, Simon},
  journal={Proceedings of the National Academy of Sciences},
  volume={111},
  number={26},
  pages={9419--9424},
  year={2014},
  publisher={National Acad Sciences}
}

@article{goh2009culture,
  title={Culture modulates eye-movements to visual novelty},
  author={Goh, Joshua O and Tan, Jiat Chow and Park, Denise C},
  journal={PLoS One},
  volume={4},
  number={12},
  pages={e8238},
  year={2009},
  publisher={Public Library of Science San Francisco, USA}
}

@article{kim2013information,
  title={Information and culture: Cultural differences in the perception and recall of information},
  author={Kim, Ji-Hyun},
  journal={Library \& information science research},
  volume={35},
  number={3},
  pages={241--250},
  year={2013},
  publisher={Elsevier}
}

@article{franceschelli2025creativity,
  title={On the creativity of large language models},
  author={Franceschelli, Giorgio and Musolesi, Mirco},
  journal={AI \& society},
  volume={40},
  number={5},
  pages={3785--3795},
  year={2025},
  publisher={Springer}
}

@book{boden2010creativity,
  title={Creativity and art: Three roads to surprise},
  author={Boden, Margaret A},
  year={2010},
  publisher={Oxford University Press}
}

@article{chua2015impact,
  title={The impact of culture on creativity: How cultural tightness and cultural distance affect global innovation crowdsourcing work},
  author={Chua, Roy YJ and Roth, Yannig and Lemoine, Jean-Fran{\c{c}}ois},
  journal={Administrative Science Quarterly},
  volume={60},
  number={2},
  pages={189--227},
  year={2015},
  publisher={Sage Publications Sage CA: Los Angeles, CA}
}

@article{zhou2017new,
  title={Is it new? Personal and contextual influences on perceptions of novelty and creativity.},
  author={Zhou, Jing and Wang, Xiaoye May and Song, Lynda Jiwen and Wu, Junfeng},
  journal={Journal of Applied Psychology},
  volume={102},
  number={2},
  pages={180},
  year={2017},
  publisher={American Psychological Association}
}

@inproceedings{alabi2024hidden,
  title={The hidden space of transformer language adapters},
  author={Alabi, Jesujoba and Mosbach, Marius and Eyal, Matan and Klakow, Dietrich and Geva, Mor},
  booktitle={Proceedings of the 62nd Annual Meeting of the Association for Computational Linguistics (Volume 1: Long Papers)},
  pages={6588--6607},
  year={2024}
}

@article{bandarkar2025multilingual,
  title={Multilingual Routing in Mixture-of-Experts},
  author={Bandarkar, Lucas and Yang, Chenyuan and Fayyaz, Mohsen and Hu, Junlin and Peng, Nanyun},
  journal={arXiv preprint arXiv:2510.04694},
  year={2025}
}

\appendix 

\section{Research Methods}

\subsection{Prompts}
\label{sec:app_prompts}

We provide two additional prompt variations used to generate the recipes with the LLMs. The first example in Figure \ref{prompt:instance_appendix1} provides a blend example in which we do not specify any country associated with the recipes. It allowed us, in section \ref{sec:trad_countries}, to analyze the mismatch between the recipe's origin country and the one proposed by LLMs when generating the recipe. The second example in Figure \ref{prompt:instance_appendix2} presents the case where the LLM is prompted as an expert of the culture of the country (origin or variation, depending on the case) before generating the recipe.

\begin{figure}
\begin{promptbox}
"Create a \textless KW\textgreater version of this recipe: \textless RECIPE-NAME\textgreater. 

Please return, in English only, the following: 
1. A recipe title. 
2. A list of ingredients. 
3. A set of cooking instructions. 

The instructions must use only the ingredients listed above, be clear and concise, and maintain the structure and order described. Title:"
\end{promptbox}
\caption{Prompt example to generate variations of a recipe given a specific keyword KW, a NATIONALITY and a RECIPE-NAME.}
\label{prompt:instance_appendix1}
\end{figure}

\begin{figure}
\begin{promptbox}
"You are knowledgeable about \textless COUNTRY\textgreater, including its culture, history, and nuances, providing insightful and context-aware responses. Create a \textless KW\textgreater version of this recipe: \textless RECIPE-NAME\textgreater.  

Please return, in English only, the following: 
1. A recipe title. 
2. A list of ingredients. 
3. A set of cooking instructions. 

The instructions must use only the ingredients listed above, be clear and concise, and maintain the structure and order described. Title:"
\end{promptbox}
\caption{Prompt example to generate variations of a recipe given a specific keyword KW, a COUNTRY name and a RECIPE-NAME.}
\label{prompt:instance_appendix2}
\end{figure}

\subsection{Metric formulation}
\label{sec:app_metric}

The central point with the novelty metrics defined in \citep{carichon2025crossing} is the definition of a cultural knowledge space that will allow to measure the divergence of a new text compared to this cultural knowledge space. The knowledge space is constituted of recipes $Recipes$ acting as artifacts of a country's culture. We also have to define a set of countries $C$ associated to the recipes. We define the set of recipe's textual instructions $T$ that are considered as the relevant depiction of the recipes. The cultural knowledge space is defined as the set of all instructions $T_{Recipe_k;c_j}$ associated with the recipe $Recipe_k$ and the country $c_j$. We define $P$ as the normal distribution, learned from this knowledge space. For each $Recipe_k$ associated with $c_j$, we have paired cultural variations, coming from different countries $c_i$, described as $VC_i = {vc_{i}^{1}, vc_{i}^{2}, ..., vc_{i}^{m}}$ each associated with the textual instructions $NT$. The new distribution $Q$, is associated with a these recipes. We can then express $P$ and $Q$ as : 

\begin{itemize}
    \item $P_{Recipe_k;c_j} = P(w|T_{Recipe_k;c_j})$ is the probability distribution of the set of documents of a given recipe $Recipe_k$ and a given country $c_j$.
    \item $Q_{Recipe_k} = P(w|NT_{Recipe_k;c_i})$ is the probability distribution of a new document $NT$ of the same recipe $Recipe_k$ and a different country $c_i$.
\end{itemize}
 
Based on these two sets, we can define the five divergence metrics.

\vspace{0.3cm}
\underline{\textit{Cultural Newness:}}
\noindent The proportion of words significantly appearing or disappearing in the distribution of the text coming from a cultural variation compared to the distribution of our cultural knowledge base.

\begin{equation}
    N_{x} = \frac{1}{N_{Q_{Recipe_k;c_i}}}\sum_{i=1}^{N_{Q_{Recipe_k;c_i}}} \delta_i \quad \text{with} \quad
    \delta_i = 
    \begin{cases}
       1 & \text{if } x_i \geq \epsilon, \\
       0 & \text{if } x_i \leq \epsilon,
    \end{cases}
\end{equation}

\noindent  Where \(N_x\) represents either \textit{Appearance} or \textit{Disappearance} depending on the choice of \( x \). $N_{Q^{ij}_{p_k}}$ is the number of words in the new text $NT_{Recipe_k}$, or the number of quantitative variables in the distribution $Q_{Recipe_k;c_i}$, and $\epsilon$ is the newness threshold. We re-use the same threshold value as in \citep{carichon2025crossing} determined through leave-one out strategy and representing the fact that a new term should contribute to the divergence more than it is observed for a cultural product associated with a specific community. The final \textit{Newness} score is :

\begin{equation}
    Newness = \lambda_1 Appearance + \lambda_2 Disappearance
\end{equation}

Where $\lambda_1 = 0.8$ and $\lambda_2 = 0.2$.

\vspace{0.3cm}
\underline{\textit{Cultural Uniqueness:}}
\noindent The divergence measure of the distribution of a cultural variation compared to the prototypical view of the cultural product within a community. 

The cultural uniqueness score is simply defined as:

\begin{equation}
    Uniqueness = D_{JS}(P_{Recipe_k;c_j}||Q_{Recipe_k;c_i})
\end{equation}

\vspace{0.3cm}
\underline{\textit{Cultural Difference:}}
\noindent The distribution of a cultural variation is distant from all the observed points in the knowledge base associated with that cultural product. 

In the neighborhood of the cultural variation recipe $VC_i$, we count the proportion of recipes that exceed a certain threshold distance. The more points there are beyond this single point, the more the recipe can be considered different from its neighborhood and, therefore, new. The formulation of the cultural difference score is as follows: 

\begin{equation}
    Difference = \frac{1}{|T_{Recipe_k;c_j}|} \sum_{m=1}^{|T_{Recipe_k;c_j}|} \delta_i with  
    \begin{cases}
       \delta_i =  1 & \text{if  }  D_{JS}(P_{t_m; Recipe_k;c_j}||Q_{Recipe_k;c_i}) \geq \epsilon, \\
       \delta_i =  0 & \text{if  }  D_{JS}(P_{t_m; Recipe_k;c_j}||Q^{ij}_{Recipe_k} \leq \epsilon, 
    \end{cases}
\end{equation}

Where $\epsilon$ is the threshold for difference and $P_{t_m; Recipe_k;c_j}$ is the probability distribution of a specific text in the set $T_{Recipe_k;c_j}$ associated with a cultural product from a community $j$. As for newness, we set the threshold as in \citep{carichon2025crossing} that considers  that a new document will be different if it is farther away than its neighbors than the community's average distance between points.

\vspace{0.3cm}
\underline{\textit{Cultural Surprise:}}
\noindent The distribution of expected combinations of attributes of a cultural variation violates the projection of the cultural knowledge space distribution into a cultural expectation base. 

\noindent This notion declines in two sub-metrics:

\noindent \textit{New Surprise:}
Given $W_T$, the set of unique $M$ words present in $T_{Recipe_k;c_j}$, associated with a recipe $k$ and a country $j$; and $W_{NT}$, the set of $N$ words in the new text $NT_{Recipe_k;c_i}$ associated with the same recipe but not necessarily the same country. Knowing that the Mutual Information between two words is defined as $MI(w_i,w_j) = \log \frac{p(w_i,w_j)}{p(w_i)p(w_j)}$, we can  create the representations of the PMI matrix $PPMI \in \mathbb{R}^{M*M}$ estimated by the co-occurrence of all terms in $W_T$ and $QPMI \in \mathbb{R}^{N*N}$ estimated from the co-occurrence of all terms in $W_{NT}$. The New Surprise score is then defined as

\begin{equation}
    New Surprise = \frac{1}{N} \sum_{i,j}^{N}
    \begin{cases}
    1 & \text{if  } QPMI(w_i,w_j) > 0 \land PPMI(w_i,w_j) =0\\
    1 & \text{if  } {w_i,w_j} \notin PPMI \\
    0 & else
    \end{cases}
\end{equation}

\noindent \textit{Divergent Surprise:}
We create $W_{TNT} = {w_1, w_2, ..., w_m}$ the intersection set between $W_T$ and $W_{NT}$ and composed of $K$ words. For each word $w_k$ in $W_{TNT}$, we can estimate the divergent surprise as:

\begin{equation}
    Divergent Surprise = \frac{1}{K} \sum_{i=1}^{K} D_{JS}(PPMI_{w_i}||QPMI_{w_i}) 
\end{equation}

\vspace{0.3cm}
\underline{\textit{Text processing and Encoding:}}
\noindent Since we want to have comparable results for estimating the correlation between novelty metrics and cultural distances for LLMs and humans, we follow the same text processing procedure described in \citep{carichon2025crossing}. More explictely, we also re-used the Spacy toolbox\footnote{https://spacy.io/} to preserve only nouns, adjectives, adverbs, numbers, and verbs in our LLMs recipes. We also also used lemmatization of produced recipes as post-processing of the LLM generation process. 

\section{Detailed results}
\label{sec:app_results}

In this section, we provide detailed results tables to interpret the graphics in the main article. The first table \ref{tab:kw_metrics} presents a detailed analysis of the divergence level based on the keyword variations used in the prompt. This table is complementary to the analysis figure \ref{fig:keywords_novel}. As mentioned in the paper, the only pattern we observe in these keyword variations is the tendency of LLMs to consistently produce newness in the notion of originality and novelty, which provides some insights into LLMs' understanding of creative definitions. 

\begin{table}[t]
\centering
\small
\setlength{\tabcolsep}{4pt}
\renewcommand{\arraystretch}{1.15}

\begin{adjustbox}{max width=\linewidth}
\begin{tabular}{llccccccccc}
\toprule
Metric & KW & gemma-2 & gemma-3 & qwen-2 & qwen-3 & llama3-70b & phi4-8b & orion-14b & falcon-40b \\
\midrule

\multirow{7}{*}{Newness}
 & novelty     & \textbf{0.027} & \textbf{0.0068} & \textbf{0.0061} & \textbf{0.0079} & \textbf{0.0104} & \textbf{0.0056} & \textbf{0.0412} & \textbf{0.0418} \\
 & uniqueness   & 0.025 & 0.0054 & 0.0047 & 0.0075 & 0.0080 & 0.0043 & 0.0342 & 0.0357 \\
 & newness     & 0.023 & 0.0053 & 0.0047 & 0.0071 & 0.0079 & 0.0043 & 0.0342 & 0.0355 \\
 & difference  & 0.024 & 0.0047 & 0.0046 & 0.0072 & 0.0078 & 0.0042 & 0.0337 & 0.0343 \\
 & surprise    & 0.024 & 0.0052 & 0.0050 & 0.0073 & 0.0082 & 0.0043 & 0.0338 & 0.0356 \\
 & creativity  & 0.024 & 0.0065 & 0.0049 & 0.0072 & 0.0086 & 0.0042 & 0.0330 & 0.0348 \\
 & originality & \textbf{0.027} & \textbf{0.0069} & \textbf{0.0058} & \textbf{0.0084} & \textbf{0.0103} & \textbf{0.0055} & \textbf{0.0449} & \textbf{0.0436} \\
\midrule

\multirow{7}{*}{Uniqueness}
 & novelty     & 0.385 & 0.474 & 0.535 & 0.541 & 0.469 & \textbf{0.535} & 0.483 & 0.402 \\
 & uniqueness  & 0.404 & \textbf{0.502} & 0.540 & 0.541 & 0.472 & 0.533 & \textbf{0.491} & 0.412 \\
 & newness     & 0.399 & 0.492 & \textbf{0.544} & \textbf{0.548} & 0.472 & 0.531 & 0.488 & 0.410 \\
 & difference  & \textbf{0.406} & 0.491 & 0.542 & 0.541 & \textbf{0.474} & 0.533 & 0.490 & \textbf{0.426} \\
 & surprise    & \textbf{0.407} & \textbf{0.501} & 0.543 & 0.532 & \textbf{0.474} & \textbf{0.539} & \textbf{0.492} & 0.413 \\
 & creativity  & 0.398 & 0.498 & \textbf{0.544} & 0.527 & 0.465 & 0.525 & 0.490 & \textbf{0.425} \\
 & originality & 0.382 & 0.462 & 0.538 & 0.539 & 0.457 & 0.527 & 0.477 & 0.390 \\
\midrule

\multirow{7}{*}{Difference}
 & novelty     & 0.714 & 0.931 & 0.995 & 0.997 & 0.961 & 0.989 & 0.943 & 0.484 \\
 & uniqueness   & \textbf{0.795} & 0.964 & 0.995 & 0.997 & \textbf{0.975} & \textbf{0.991} & \textbf{0.956} & 0.539 \\
 & newness     & 0.773 & 0.957 & 0.997 & 0.998 & 0.971 & 0.990 & 0.953 & 0.528 \\
 & difference  & 0.794 & 0.956 & 0.995 & 0.997 & 0.971 & 0.989 & 0.953 & \textbf{0.562} \\
 & surprise    & \textbf{0.823} & \textbf{0.969} & 0.995 & 0.996 & 0.974 & \textbf{0.991} & 0.953 & 0.537 \\
 & creativity  & 0.772 & \textbf{0.972} & 0.998 & 0.997 & \textbf{0.976} & 0.988 & 0.953 & \textbf{0.554} \\
 & originality & 0.659 & 0.908 & 0.995 & 0.997 & 0.947 & 0.987 & 0.936 & 0.469 \\
\midrule

\multirow{7}{*}{New Surprise}
 & novelty     & \textbf{0.766} & 0.900 & 0.966 & 0.972 & \textbf{0.912} & \textbf{0.952} & 0.811 & \textbf{0.499} \\
 & uniqueness  & 0.754 & 0.896 & 0.967 & 0.971 & 0.904 & \textbf{0.949} & 0.810 & 0.481 \\
 & newness     & \textbf{0.768} & \textbf{0.914} & 0.968 & 0.972 & \textbf{0.910} & 0.947 & 0.816 & 0.488 \\
 & difference  & 0.730 & 0.861 & 0.967 & 0.970 & 0.902 & 0.944 & \textbf{0.824} & 0.459 \\
 & surprise    & 0.734 & 0.867 & 0.967 & 0.971 & 0.898 & 0.945 & \textbf{0.819} & 0.477 \\
 & creativity  & 0.750 & 0.900 & 0.966 & 0.969 & 0.899 & 0.942 & 0.817 & 0.485 \\
 & originality & 0.751 & 0.886 & 0.965 & 0.972 & 0.901 & 0.941 & 0.808 & \textbf{0.538} \\
\midrule

\multirow{7}{*}{DivergentSurprise}
 & novelty     & \textbf{0.483} & 0.514 & 0.542 & 0.549 & 0.529 & 0.526 & 0.447 & \textbf{0.348} \\
 & uniqueness  & \textbf{0.481} & \textbf{0.515} & 0.542 & \textbf{0.550} & \textbf{0.532} & 0.525 & 0.458 & 0.338 \\
 & newness     & 0.480 & 0.514 & 0.538 & 0.548 & 0.531 & 0.519 & 0.458 & 0.338 \\
 & difference  & 0.466 & 0.500 & 0.543 & 0.551 & 0.530 & 0.524 & \textbf{0.463} & 0.322 \\
 & surprise    & 0.470 & 0.509 & 0.542 & 0.552 & 0.528 & \textbf{0.529} & 0.458 & 0.331 \\
 & creativity  & 0.476 & 0.513 & 0.537 & \textbf{0.559} & \textbf{0.533} & 0.527 & 0.456 & 0.334 \\
 & originality & 0.476 & 0.509 & 0.536 & 0.548 & 0.531 & 0.527 & 0.450 & \textbf{0.372} \\
\bottomrule
\end{tabular}
\end{adjustbox}

\caption{Divergence values for each metric and for each models based on the ''creative'' keywords used in the prompt. Results in \textbf{bold} highlight the highest values.}
\label{tab:kw_metrics}
\end{table}

The second detailed table \ref{tab:regional_ingr_overlap} presents the regional statistic of overlap and preservation of ingredients as highlighted in section \ref{sec:mat_ground} complementing the general overlap results introduced in \ref{fig:ingr_overlap}.


\begin{table}[t]
\centering
\small
\setlength{\tabcolsep}{3pt}
\renewcommand{\arraystretch}{1.15}

\begin{adjustbox}{max width=\linewidth}
\begin{tabular}{l *{8}{cc}}
\toprule
\multirow{2}{*}{Continent}
& \multicolumn{2}{c}{gemma-2}
& \multicolumn{2}{c}{gemma-3}
& \multicolumn{2}{c}{qwen-2}
& \multicolumn{2}{c}{qwen-3}
& \multicolumn{2}{c}{llama3-70b}
& \multicolumn{2}{c}{phi4-8b}
& \multicolumn{2}{c}{orion-14b}
& \multicolumn{2}{c}{falcon-40b}
\\
\cmidrule(lr){2-3}\cmidrule(lr){4-5}\cmidrule(lr){6-7}\cmidrule(lr){8-9}
\cmidrule(lr){10-11}\cmidrule(lr){12-13}\cmidrule(lr){14-15}\cmidrule(lr){16-17}
& Overlap & Preserv.
& Overlap & Preserv.
& Overlap & Preserv.
& Overlap & Preserv.
& Overlap & Preserv.
& Overlap & Preserv.
& Overlap & Preserv.
& Overlap & Preserv.
\\
\midrule

Asia
& 80.25 & 84.10
& 92.98 & 81.65
& 91.52 & 82.03
& 85.89 & 82.73
& 87.80 & 82.38
& 88.53 & 84.20
& 76.37 & 92.96
& 62.70 & 88.75
\\

Europe
& 80.54 & 79.54
& 93.58 & 77.80
& 91.96 & 77.67
& 86.36 & 77.83
& 86.90 & 77.85
& 88.31 & 80.45
& 70.60 & 94.14
& 55.87 & 86.18
\\

North America
& 78.24 & 83.75
& 92.31 & 81.27
& 91.19 & 80.26
& 85.86 & 80.82
& 86.94 & 80.73
& 88.05 & 82.58
& 71.75 & 95.70
& 54.70 & 89.92
\\

Oceania
& 85.88 & 77.77
& 95.83 & 79.62
& 94.68 & 81.48
& 92.10 & 83.33
& 94.10 & 81.48
& 97.18 & 79.62
& --    & --
& 72.00 & 87.03
\\

South America
& 86.81 & 73.49
& 95.87 & 72.91
& 94.69 & 72.24
& 90.73 & 72.52
& 90.01 & 71.74
& 92.64 & 75.21
& --    & --
& 64.39 & 79.44
\\

Carribean
& 80.84 & 86.60
& 92.67 & 84.27
& 93.58 & 83.18
& 87.51 & 83.52
& 86.27 & 85.37
& 88.18 & 86.08
& 88.46 & 97.05
& 57.32 & 89.88
\\

Middle East
& 85.02 & 71.58
& 95.28 & 70.54
& 94.77 & 66.87
& 87.61 & 68.91
& 90.80 & 68.96
& 90.04 & 72.38
& 77.85 & 86.32
& 64.34 & 79.81
\\

Africa
& 81.82 & 81.67
& 93.64 & 79.15
& 92.17 & 77.27
& 85.96 & 78.07
& 90.80 & 77.70
& 88.86 & 80.50
& 88.88 & 96.02
& 63.68 & 88.87
\\

\bottomrule
\end{tabular}
\end{adjustbox}

\caption{Per-continent ingredient overlap (ingredients used in references) and preservation (ingredients from references retained in recipes) for each model.}
\label{tab:regional_ingr_overlap}
\end{table}

Finally, we provide the top ingredients used by LLMs in their generated recipes in Table \ref{tab:top20_ingredients}. We indicate the total number of recipes that include these ingredients. For example, among the 1M300K recipes generated by our eight models, 392K (30\%) contain the keyword ''Salt taste'' in their ingredient lists.

\begin{table}[t]
\centering
\small
\setlength{\tabcolsep}{6pt}
\renewcommand{\arraystretch}{1.15}

\begin{tabular}{lr}
\hline
Ingredient & Total count \\
\hline
Salt taste & 392,342 \\
Onion & 356,069 \\
Garlic & 302,287 \\
Salt & 280,364 \\
Oil & 201,222 \\
Sugar & 180,278 \\
Pepper taste & 175,503 \\
Butter & 164,215 \\
Purpose flour & 106,931 \\
Pepper & 81,077 \\
Optional & 74,294 \\
Egg & 59,370 \\
Water & 52,447 \\
Milk & 38,705 \\
Powder & 38,047 \\
Oil frying & 34,682 \\
Parsley & 34,164 \\
Taste & 31,721 \\
Room temperature & 24,365 \\
Vanilla extract & 22,283 \\
\hline
\end{tabular}

\caption{Top 20 ingredients ranked by total frequency across all datasets.}
\label{tab:top20_ingredients}
\end{table}

\end{document}